\documentclass{article}

\usepackage{microtype}
\usepackage{graphicx}
\usepackage{subfigure}
\usepackage{booktabs} % for professional tables

\usepackage{hyperref}
\usepackage{caption}

\usepackage[accepted]{icml2025}

\usepackage{amsmath}
\usepackage{amssymb}
\usepackage{mathtools}
\usepackage{amsthm}
\usepackage{multirow}

\usepackage[capitalize,noabbrev]{cleveref}

\theoremstyle{plain}

\theoremstyle{definition}

\theoremstyle{remark}

\usepackage[textsize=tiny]{todonotes}

\newcommand{\Real}{\mathbb{R}}
\newcommand\vect[1]{\pmb{#1}}
\newcommand\vectcal[1]{\pmb{\mathcal{#1}}}
\DeclarePairedDelimiter\norm{\lVert}{\rVert}
\DeclarePairedDelimiterX\innerp[2]{\langle}{\rangle}{#1
  \mathop{}\delimsize\vert\mathop{} #2}

\Crefname{equation}{Eq.}{Eqs.}

\DeclarePairedDelimiterX\Set[1]{\lbrace}{\rbrace}%
{  #1 }

\DeclareMathOperator{\avgpool}{AvgPool}
\DeclareMathOperator{\quant}{Q}

\makeatletter
\newcommand*{\ie}{%
  \@ifnextchar{,}%
  {i.e.}%
  {i.e.,\@\xspace}%
}
\newcommand*{\eg}{%
  \@ifnextchar{,}%
  {e.g.}%
  {e.g.,\@\xspace}%
}
\newcommand*{\cf}{%
  \@ifnextchar{,}%
  {cf.}%
  {cf.,\@\xspace}%
}
\makeatother

\allowdisplaybreaks

\icmltitlerunning{Model-Free Adversarial Purification via Coarse-To-Fine Tensor Network Representation}

\begin{document}

\twocolumn[
\icmltitle{Model-Free Adversarial Purification via Coarse-To-Fine Tensor Network Representation}
% Training-Free Tensor Defender

% Training-Free Tensor Defender: Advancing Accuray, Robustness and Generalization

% Tensor Decomposition Defense: A Training-Free Defense Method for Adversarial Attacks

% Training-Free Tensor Decomposition for Adversarial Purification 

% XXX Tensor Decomposition for Adversarial Purification

% Training-Free Tensor Defender: Advancing Robustness and Generalization via Tensor Low-rank Prior



\icmlsetsymbol{equal}{$\dagger$}
\icmlsetsymbol{corrauthor}{*}

\begin{icmlauthorlist}
\icmlauthor{Guang Lin}{equal,1,2}
\icmlauthor{Duc Thien Nguyen}{equal,1,3}
\icmlauthor{Zerui Tao}{1}
\icmlauthor{Konstantinos Slavakis}{3}
\icmlauthor{Toshihisa Tanaka}{1,2}
\icmlauthor{Qibin Zhao}{corrauthor,1,2}
\end{icmlauthorlist}

\icmlaffiliation{1}{RIKEN Center for Advanced Intelligence Project (RIKEN AIP) }
\icmlaffiliation{2}{Tokyo University of Agriculture and Technology }
\icmlaffiliation{3}{Institute of Science Tokyo }

\icmlcorrespondingauthor{}{}

\icmlkeywords{Trustworthy Machine Learning, ICML}

\vskip 0.3in
]



\printAffiliationsAndNotice{\icmlEqualContribution} % otherwise use the standard text.

\begin{abstract}

Deep neural networks are known to be vulnerable to well-designed adversarial attacks. Although numerous defense strategies have been proposed, many are tailored to the specific attacks or tasks and often fail to generalize across diverse scenarios.
In this paper, we propose Tensor Network Purification (TNP), a novel model-free adversarial purification method by a specially designed tensor network decomposition algorithm. TNP depends neither on the pre-trained generative model nor the specific dataset, resulting in strong robustness across diverse adversarial scenarios.
To this end, the key challenge lies in relaxing Gaussian-noise assumptions of classical decompositions and accommodating the unknown distribution of adversarial perturbations. Unlike the low-rank representation of classical decompositions, TNP aims to reconstruct the unobserved clean examples from an adversarial example. Specifically, TNP leverages progressive downsampling and introduces a novel adversarial optimization objective to address the challenge of minimizing reconstruction error but without inadvertently restoring adversarial perturbations.
Extensive experiments conducted on CIFAR-10, CIFAR-100, and ImageNet demonstrate that our method generalizes effectively across various norm threats, attack types, and tasks, providing a versatile and promising adversarial purification technique.

\end{abstract}

\section{Introduction}
\label{Introduction}
Deep neural networks (DNNs) have achieved remarkable success across a wide range of tasks. However, DNNs have been shown to be vulnerable to adversarial examples \citep{szegedy2013intriguing,goodfellow2014explaining}, which are generated by adding small, human-imperceptible perturbations to natural images but completely change the prediction results to DNNs, leading to disastrous implications. The vulnerability of DNNs to such examples highlights the significance of robust defense mechanisms to mitigate adversarial attacks effectively.

Since then, numerous methods have been proposed to defend against adversarial examples.
Notably, adversarial training \citep[AT,][]{goodfellow2014explaining} typically aims to retrain DNNs using adversarial examples, achieving robustness over seen types of adversarial attacks but performing poorly against unseen perturbations \citep{laidlaw2021perceptual,dolatabadi2022}.
Another class of defense methods is adversarial purification \citep[AP,][]{yoon2021adversarial}, which leverages pre-trained generative models to remove adversarial perturbations and demonstrates better generalization than AT against unseen attacks \citep{nie2022diffusion,lin2024adversarial}. However, AP methods rely on pre-trained models tailored to specific datasets, limiting their transferability to different data distributions and tasks.
Thus, both mainstream techniques face generalization challenges: AT struggles with diverse norm threats, and AP with task generalization, restricting their applicability to broader scenarios.

To tackle these challenges, we propose a novel model-free adversarial purification method by a specially designed tensor network decomposition algorithm, termed Tensor Network Purification (TNP), which bridges the gap between low-rank tensor network representation with Gaussian noise assumption and removal of adversarial perturbations with unknown distributions.
As a model-free optimization technique \citep{oseledets2011tensor,zhao2016tensor}, TNP depends neither on any pre-trained generative model nor specific dataset, enabling it to achieve strong robustness across diverse adversarial scenarios.
Additionally, TNP can eliminate potential adversarial perturbations for both clean or adversarial examples before feeding them into the classifier \citep{yoon2021adversarial}. As a pre-processing step, TNP can defend against adversarial attacks without retraining the classifier model or pretraining the generative model. Moreover, since our method is an algorithm applied to input examples only and has no fixed model parameters, it is more difficult to be attacked by existing black or white box adversarial attack techniques.
Consequently, benefiting from the aforementioned advantages, it is evident that TN-based AP methods represent a highly promising direction, offering the transferability to be effectively applied across a wide range of adversarial scenarios.

The existing TN methods are particularly favorable for image completion and denoising when the noise is sparse or follows Gaussian distribution as long as it can be modeled explicitly.
However, the distribution of well-designed adversarial perturbations fundamentally differs from such noises and often aligns with the statistics of the data \citep{ilyas2019adversarial,allen2022feature}.
Consequently, these perturbations behave more like features than noise, making them challenging to be modeled explicitly and prone to being inadvertently reconstructed.
To address this issue, we first explore the distribution changes of perturbations during the optimization process and initially mitigate the impact of perturbations by progressive downsampling. 
Correspondingly, we propose an algorithm for TN incremental learning in a coarse-to-fine manner. Furthermore, a novel adversarial optimization objective is proposed to address the challenge of minimizing reconstruction error but without inadvertently restoring adversarial perturbations.
Unlike classical TN, given an adversarial example, TNP prevents naive low-rank representation of the input and encourages the reconstructed examples to approximate the unobserved clean examples.

We empirically evaluate the performance of our method by comparing it with AT and AP methods across attack settings using multiple classifier models on CIFAR-10, CIFAR-100, and ImageNet.
The results demonstrate that our method achieves state-of-the-art performance with robust generalization across diverse adversarial scenarios.
Specifically, our method achieved a 26.45\% improvement in average robust accuracy over AT across different norm threats, a 9.39\% improvement over AP across multiple attacks, and a 6.47\% improvement over AP across different datasets.
Furthermore, in denoising tasks, our method effectively removes adversarial perturbations while preserving consistency between the reconstructed clean example and the reconstructed adversarial example.
These results collectively underscore the effectiveness and potential of our proposed method.
In summary, our contributions are as follows:
\begin{itemize}
    \item We introduce a model-free adversarial purification framework based on tensor network representation, which eliminates the need for training a powerful generative model or relying on specific dataset distributions, making it a general-purpose adversarial purification.
    \item Based on our analysis of the distribution changes of adversarial perturbations during optimization, we propose a novel adversarial optimization objective for coarse-to-fine TN representation learning to prevent the restoration of adversarial perturbations.
    \item We conduct extensive experiments on various datasets, demonstrating that our method achieves state-of-the-art performance, especially exhibiting robust generalization across diverse adversarial scenarios.
\end{itemize}

\begin{figure*}[t]
\vskip 0.2in
    \centering
    \includegraphics[width=\linewidth]{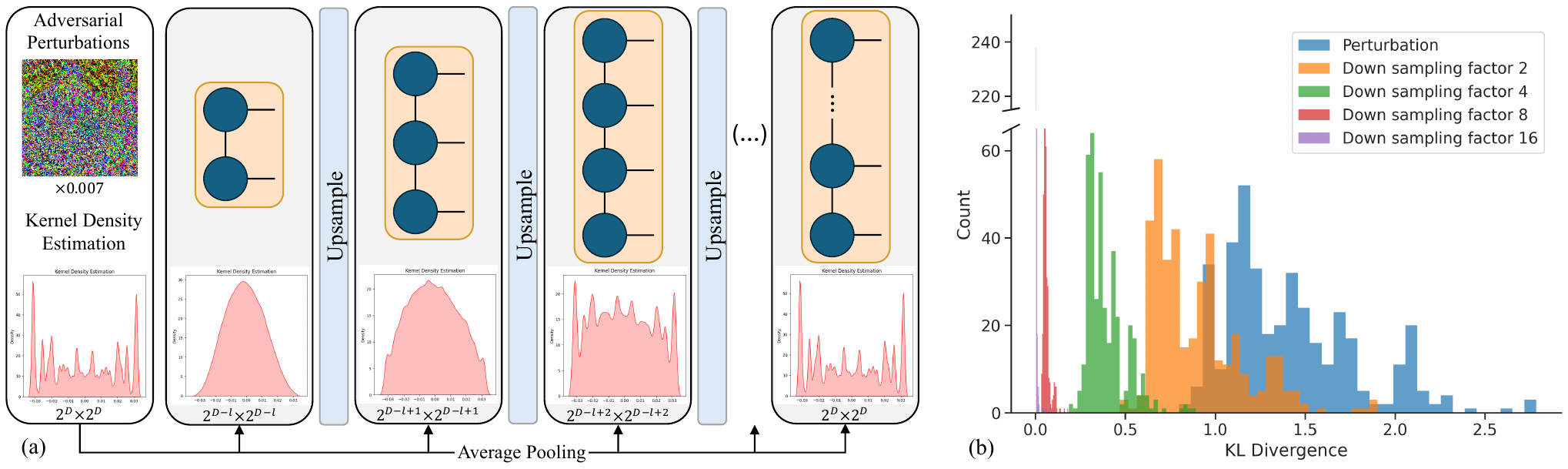}
    \caption{Compare the adversarial perturbations in the downsampled images. (a) The distribution changes of adversarial perturbations during downsampling process. More results are shown in \Cref{app:distribution}. (b) The KL divergence histogram of adversarial perturbations.}
    \label{fig:distribution}
    \vskip -0.1in
\end{figure*}

\section{Related Works}
\label{Related}
\textbf{Adversarial robustness} \quad To defend against adversarial attacks, researchers have developed various techniques aimed at enhancing the robustness of DNNs.
\citet{goodfellow2014explaining} propose AT technique to defend against adversarial attacks by retraining classifiers using adversarial examples. In contrast, AP methods \citep{shi2021online,srinivasan2021robustifying} aim to purify adversarial examples before classification without retraining the classifier. Currently, the most common AP methods \citep{nie2022diffusion,bai2024diffusion} rely on pre-trained generative models as purifiers, which are trained on specific datasets and hard to generalize to data distributions outside their training domain. \citet{lin2024adversarial} propose applying AT technique to AP, optimizing the purifier to adapt to new data distributions.
Although our framework employs AP strategy, it fundamentally deviates from these works as we develop a model-free framework that relies solely on the information of the input example for adversarial purification, without requiring any additional priors from pre-trained models or training the purifiers.

\textbf{Tensor network and TN-based defense methods} \quad Tensor network (TN) is a classical tool in signal processing, with many successful applications in image completion and denoising \citep{kolda2009tensor,cichocki2015tensor}.
Compared to traditional TN methods such as TT \citep{oseledets2011tensor} and TR \citep{zhao2016tensor}, we employ the quantized technique \citep{khoromskij2011d} and develop a coarse-to-fine strategy. Recently, PuTT \citep{loeschcke2024coarse} also employs a coarse-to-fine strategy, aiming to achieve better initialization for faster and more efficient TT decomposition by minimizing the reconstruction error.
In comparison, our method progresses from low to high resolution, explicitly targeting perturbation removal and analyzing the impact of downsampling on perturbations. Furthermore, we propose a novel optimization objective that goes beyond simply minimizing the reconstruction error, focusing instead on preventing the reappearance of adversarial perturbations.

With the growing concern over adversarial robustness, a line of work has attempted to leverage TNs as robust denoisers to defend against adversarial attacks. In particular, \citet{yang2019me} reconstruct images and retrain classifiers to adapt to the new reconstructed distribution. \citet{entezari2022tensorshield} analyze vanilla TNs and show their effectiveness in removing high-frequency perturbations.
Additionally, \citep{bhattarai2023robust} extend the application of TNs beyond data to include classifiers, a concept similar to the approaches of \citep{rudkiewicz2024robustness,phan2023cstar}. Furthermore, \citep{song2024training} employ training-free techniques while incorporating ground truth information to defend against adversarial attacks. However, the aforementioned methods rely on additional prior or are limited to specific attacks.
In this paper, we aim to achieve robustness solely by optimizing TNs themselves, establishing them as a plug-and-play and promising adversarial purification technique.

\section{Backgrounds}
\label{Backgrounds}

\textbf{Notations} \quad Throughout the paper, we denote scalars, vectors, matrices, and tensors as lowercase letters, bold lowercase letters, bold capital letters, and calligraphic bold capital letters, \eg, $x$, $\vect{x}$, $\vect{X}$ and $\vectcal{X}$, respectively.
A $D$-order tensor tensor is an $D$-dimensional array, \eg, a vector is a 1st-order tensor and a matrix is a 2nd-order tensor.
For a $D$-order tensor $\pmb{\mathcal{X}} \in \mathbb{R}^{I_1 \times \dots \times I_D}$, we denote its ($i_1,\dots,i_D$)-th entry as $x_{\mathbf{i}}$, where $\mathbf{i} =$ ($i_1,\dots,i_D$).
% For a positive integer $n$, denote $[n] \coloneqq {1,2,\ldots, n}$.
Following the conventions in deep learning, we treat images as vectors, \eg, input example $\vect{x}_{in}$, clean example $\vect{x}_{cln}$, adversarial example $\vect{x}_{adv}$ and reconstructed example $\vect{y}$.

\textbf{Tensor network decomposition} \quad
Given a $D$-order tensor $\vectcal{X} \in \Real^{I_1 \times \ldots \times I_D}$,
tensor network decomposition factorizes $\vectcal{X}$ into $D$ smaller latent components
by using some predefined tensor contraction rules.
% The classical Tucker deomposition \citep{tucker1966some} assumes
% \(
% \vectcal{X}=\vectcal{W} \times_1 \vect{Z}^1 \times_2 \ldots \times_D \vect{Z}^D \,,
% \)
% where \(
% \vectcal{W} \in \Real^{R_1 \times \ldots \times R_D} \,, \vect{Z}^d \in \Real^{I_d\times R_d} \,,
% \forall d \in [D] \,,
% \)
% and $\times_d$ dnotes the matrix-tensor contraction \citep{kolda2009tensor}.
% Equivalently, each entry can be written as
% \(
% x_{\mathbf{i}} = \sum_{r_1=1}^{R_1} \cdots \sum_{r_D=1}^{R_D} w_{r_1\ldots r_D} z^1_{i_1 r_1}
% \cdots z^D_{i_D r_D} \,,
% \)
% where the tuple $(R_1,\ldots, R_D)$ is the Tucker rank of $\vectcal{X}$.
% The latent factors $\vect{Z}_d$ can capture information of each tensor mode and
% $\vectcal{W}$ represents the weight of each factors.
% CP decomposition \citep{hitchcock1927expression} is a special case of Tucker when $\vectcal{W}$ is super-diagonal, \ie,
% \(
% x_{\mathbf{i}} = \sum_{r=1}^{R} w_{r} z^1_{i_1 r} \cdots z^D_{i_D r} \,,
% \)
% and $R$ is the CP rank of $\vectcal{X}$.
% Tensor Train (TT) decomposition \citep{oseledets2011tensor} improves the approximation guarantee of CP and the compression
% rate of Tucker decompositions.
Among tensor network decompositions, Tensor Train (TT) decomposition \citep{oseledets2011tensor} enjoys both quasi-optimal approximation as well as the high compression rate of large and complex data tensors.
In particular, a $D$-order tensor $\vectcal{X} \in \Real^{I_1\times \ldots \times I_D}$ has the TT format as
\(
{x}_{\mathbf{i}}= \vect{A}^1_{i_1} \vect{A}^2_{i_2} \ldots \vect{A}^D_{i_D} \,,
\)
where $\vect{A}^d_{i_d} \in \Real^{r_{d-1}\times r_d}$, for $d\in[D]$ and $i_d \in [I_d]$. Then, $(1,r_1,\ldots,r_{d-1},1)$ is the TT rank of $\vectcal{X}$.
For simplicity, we denote $\vectcal{X}=\text{TT}(\vectcal{A}^1, \ldots, \vectcal{A}^D)$.
When each dimension $I_d$ of $\vectcal{X}$ is large,
quantized tensor train \citep[QTT,][]{khoromskij2011d} becomes highly efficient, which
splits each dimension in powers of two.
For example, a $2^D \times 2^D$ image can be rearranged into a more expressive and balanced
$D$-order tensor.
For brevity, hereafter, a $2^D \times 2^D$ image $\vect{x}_D$ shall be called a resolution $D$ image, whose quantized tensor is $\vectcal{X}_D = \quant(\vect{x}_D)$.

\section{Method}
\label{Method}

Tensor network (TN) is a classical tool in signal processing, with many successful applications in image completion and denoising. By leveraging the $\ell_2$-norm as the primary optimization criterion, which aligns well with the statistical properties of a normal distribution, these methods \citep{phan2020tensor,loeschcke2024coarse} have demonstrated strong capabilities in removing Gaussian noise.

However, the distribution of well-designed adversarial perturbations is essentially different from Gaussian noise and cannot be modeled explicitly \citep{ilyas2019adversarial,allen2022feature}, thereby challenging the conventional assumptions of TN-based denoising methods, leading to ineffectiveness on adversarial purification for $\vect{x}_{adv}$.
To minimize the loss $\|\vect{x}_{adv} - \text{TN}(\vect{x}_{adv})\|_{2}$, TN decompositions fit all feature components of $\vect{x}_{adv}$, including the adversarial perturbations. However, in the presence of adversarial attacks, we aim to restore unobserved $\vect{x}_{cln}$ from the input $\vect{x}_{adv}$, that is: $\text{TN}(\vect{x}_{adv}) \approx \vect{x}_{cln}$ rather than $\vect{x}_{adv}$.

Based on the above analysis, it is crucial to overcome two challenges in designing an effective TN method:

\quad \emph{Q1. How can we transform the distribution of non-specific perturbations into well-known distributions?}

\quad \emph{Q2. How can we avoid overly constraints of reconstruction error from inadvertently restoring those perturbations?}

% Q2. How can we maintain the balance between minimizing the reconstruction error and removing perturbations?

To address these two issues, we explore how adversarial perturbations evolve when using average pooling as downsampling. Intuitively, the central limit theorem suggests that as an image is progressively downsampled, aggregated perturbations begin to resemble a normal distribution. Thus, even an $\ell_2$-based penalty becomes effective in suppressing the perturbations at lower resolutions.

However, while this insight helps suppress perturbations at lower resolutions, there remains the challenge of reconstructing the full-resolution image. When upsampling and further optimizing using $\|\vect{x}_{adv} - \text{TN}(\vect{x}_{adv})\|_{2}$, the perturbations will still be restored. This connects with our second question, for which we design a new optimization objective.

\subsection{Downsampling using average pooling}

An intuitive explanation for why downsampling aids in perturbation removal can be derived from the Central Limit Theorem \citep[CLT,][]{grzenda2008conditional}. When an image is downsampled by average pooling, the random components (e.g., pixel-level noise or minor adversarial perturbations) within those pooling patches are aggregated.
We hypothesize that, given an adversarial example $\vect{x}_{adv}$,  downsampling the $\vect{x}_{adv}$ from its original resolution $D$ to a lower resolution~$D-1$ will smooth out the adversarial perturbations. 
As the downsampling process progresses further, the distribution of the aggregated adversarial perturbations in the low-resolution image $\vect{x}_{D-l}$ is expected to converge toward a normal distribution, as illustrated in \Cref{fig:distribution}a.

To investigate this hypothesis in real datasets, we compute the KL divergence between the histograms of adversarial perturbations and the Gaussian distributions with the same sample mean and variance across 512 images from ImageNet. As shown in \Cref{fig:distribution}b, the distribution of those perturbations progressively aligns with that of Gaussian noise as the downsampling process progresses. Additionally, to further support our hypothesis of using the average pooling, we compare the influence of different downsampling methods, as discussed in \Cref{app:avgpool}.
% Consequently, tensor network decomposition methods can effectively suppress or remove perturbations in the low-resolution image.

% The following theorem provides a theoretical confirms that the noisy clean example distribution and the adversarial example distribution get closer after average pooling process, indicating that the downsampling process can indeed mitigate the impact of perturbations.

% \textbf{Theorem 1.} \gl{If we still have time.}

\subsection{Tensor network purification}

\begin{figure}[t]
\vskip 0.1in
    \centering
    \includegraphics[width=\linewidth]{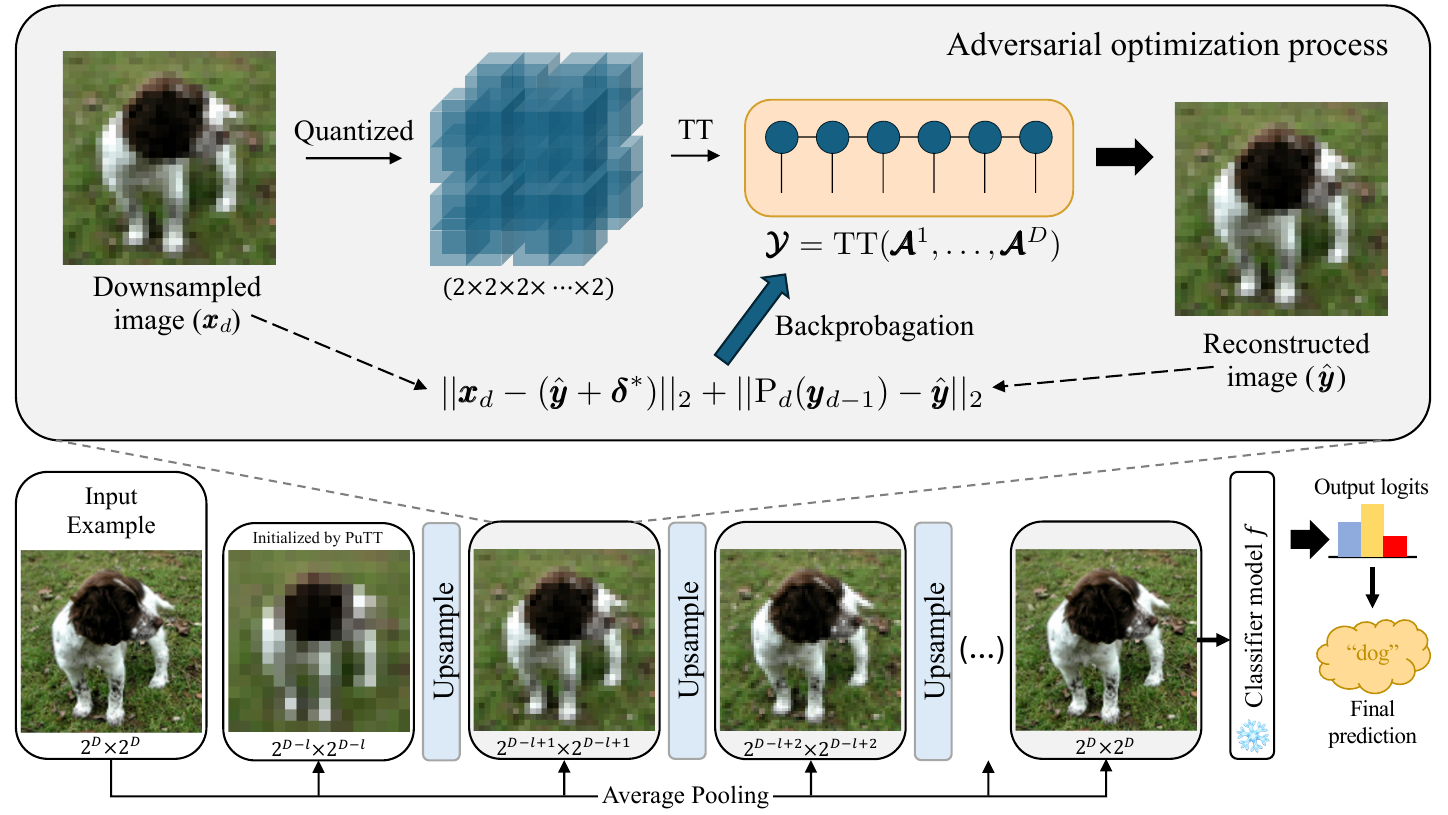}
    \caption{Illustration of tensor network purification.}
    \label{fig:process}
    \vskip -0.2in
\end{figure}

Building upon our downsampling-based intuition, we
design a coarse-to-fine process
and adopt PuTT \citep{loeschcke2024coarse} as our base model, which
employs progressive
downsampling for better initialization of QTT cores.
The workflow of tensor network purification (TNP) for classification tasks is illustrated in \Cref{fig:process}, where the quantized $\vectcal{X} = \quant(\vect{x})$, TT decomposition $\vectcal{X} \approx \vectcal{Y}=\text{TT}(\vectcal{A}^1, \ldots, \vectcal{A}^D)$, and reconstruction $\vect{y} = \quant^{-1}(\vectcal{Y})$ processes are depicted.

Initially, the $2^D \times 2^D$ input example $\vect{x}_D$ (potentially adversarial example $\vect{x}_{adv}$ or clean example $\vect{x}_{cln}$),
whose quantized version is a $D$-order tensor $\vectcal{X}_{D}$, is first downsampled to a resolution $D-l$ example
$\vect{x}_{D-l}$,
corresponding to a $(D-l)$-order tensor $\vectcal{X}_{D-l}$.
The QTT cores of $\vectcal{X}_{D-l}$ are optimized by PuTT via backpropagation within a standard reconstruction error  $||\vect{x}_{D-l}-\vect{y}_{D-l}||_2$.
Once the approximation of $\vectcal{X}_{D-l}$ is stabilized,
the prolongation operator $\vectcal{P}_{D-l+1}$ is applied to
the QTT format of $\vectcal{X}_{D-l}$, producing a $(D-l+1)$-order tensor
$\vectcal{P}_{D-l+1}\vectcal{X}_{D-l}$.
Additionally, we define the linear function $\text{P}_d(\cdot)$ acts on the image level, with the effect of upsampling from resolution $d-1$ to $d$,
details in \cref{app:TN.P}.
This serves as an initialization to find the optimal QTT cores of $\vectcal{X}_{D-l+1}$ and reconstructed downsampled example $\vect{y}_{D-l}$.

Next, the input example $\vect{x}_D$ is once again downsampled to a resolution $D-l+1$ example $\vect{x}_{D-l+1}$. However, this time, the QTT cores of $\vectcal{X}_{D-l+1}$ are optimized using the adversarial optimization objective within a novel loss function as shown in \Cref{eq:ours}. Similarly, once the approximation of $\vectcal{X}_{D-l+1}$ stabilizes, the upsampling operation is performed. This process is repeated iteratively until reaching the QTT approximation $\vectcal{Y}_D$ of the original resolution $\vectcal{X}_{D}$.

Finally, tensor network purification (TNP) can purify the potential adversarial examples ($\vect{x}_{cln}$ or $\vect{x}_{adv}$) before feeding them into the classifier model $f$, \eg, $f(\text{TNP}(\vect{x}_{cln}))=f(\text{TNP}(\vect{x}_{adv}))=y$, where $y$ is the ground truth. As a plug-and-play module, TNP can be integrated with any classifiers.

\subsection{Adversarial optimization process}
\label{Adversarial optimization process}

Following the coarse-to-fine process, despite the downsampling with average pooling and subsequent PuTT at lower resolutions can mitigate adversarial perturbations, the other challenge arises upon reconstructing the example at the original resolution, where minimizing standard reconstruction error will inevitably restore those perturbations.

\begin{table}[t]
\centering
\vskip -0.1in
\begin{minipage}{\linewidth}
\begin{algorithm}[H]
    \caption{Adversarial optimization process.}
    \label{alg:ours}
    \begin{algorithmic}
        \STATE \textbf{Input:} Example $\vect{x}_d$, number of iterations $T$, inner maximization steps $N$, scale $\alpha$ and $\eta$, learning rate $\beta$
        \STATE \textbf{Output:} Reconstructed examples $\vect{y}_d$
        \STATE Initialize $\hat{\vect{y}} \gets \text{P}_{d}(\vect{y}_{d-1})$
        \FOR{$t=1, 2, \dots, T$}
        \STATE Initialize \( \boldsymbol{\delta} \leftarrow \vect{0} \)
        \FOR{$n=1, 2, \dots, N$}
        \STATE $\ell \gets \mathcal{L}_{adv}(\hat{\vect{y}} + \boldsymbol{\delta}, \vect{x}_d)$
        \STATE $\boldsymbol{\delta} \gets \text{clip}(\boldsymbol{\delta} + \alpha\text{sign}(\nabla_{\hat{\vect{y}}}\ell), -\eta, \eta)$
        \ENDFOR
        \STATE $\boldsymbol{\delta}^* \gets \text{clip}(\hat{\vect{y}}+\boldsymbol{\delta}, 0, 1)-\hat{\vect{y}}$
        \STATE Gradient descent based on loss in \Cref{eq:ours}:
        % \STATE Update $\hat{\vect{y}}$ by $\arg\min_{\hat{\vect{y}}} \mathcal{L}(\vect{x}_d, \hat{\vect{y}}, \boldsymbol{\delta}^*)$ with \Cref{eq:ours}.
        % \STATE $\hat{\vect{y}} \gets \arg\min_{\hat{\vect{y}}} \mathcal{L}(\vect{x}_d, \hat{\vect{y}}, \boldsymbol{\delta}^*)$
        \STATE $\hat{\vect{y}} \gets \hat{\vect{y}} - \beta\nabla_{\hat{\vect{y}}}\mathcal{L}$
        \ENDFOR
        \STATE \textbf{return} $\vect{y}_d \gets \hat{\vect{y}}$
    \end{algorithmic}
\end{algorithm}
\end{minipage}
\vskip -0.2in
\end{table}

Unlike traditional reconstruction tasks, in the context of adversarial attacks, we can only observe the adversarial example $\vect{x}_{adv}$, while the goal is to reconstruct a ``clean'' version $\vect{y}$ closing to the unobserved clean example $\vect{x}_{cln}$. To bridge the gap between $\vect{x}_{adv}$ and $\vect{x}_{cln}$, we propose a new optimization objective that introduces an auxiliary variable $\boldsymbol{\delta}$ by inner maximization. Additionally, we leverage a reconstructed downsampled example as a crucial prior to guide the approximation toward $\vect{x}_{cln}$.

Here, we outline the optimization procedure for $x_d$, which corresponds to the gray box in \Cref{fig:process}.
Formally, given the resolution~$d$ example $\vect{x}_d$, we attempt to obtain the reconstructed example $\vect{y}_d$ by performing gradient descent on optimization loss functions of
\begin{equation}
\label{eq:ours}
\begin{aligned}
    \mathcal{L}(\vect{x}_d, \hat{\vect{y}}, \boldsymbol{\delta}^*) &= ||\vect{x}_d - (\hat{\vect{y}}+\boldsymbol{\delta}^*)||_2 
    + ||\text{P}_{d} (\vect{y}_{d-1}) - \hat{\vect{y}}||_2, \\
    \text{s.t.  } \boldsymbol{\delta}^* &= \arg\max_{\norm{\boldsymbol{\delta}} < \eta} \mathcal{L}_{adv}(\hat{\vect{y}} + \boldsymbol{\delta}, \vect{x}_d),
\end{aligned}
\end{equation}
where $d \in [D-l+1, D]$ and $\eta$ is a scale hyperparameter.

The auxiliary variable $\boldsymbol{\delta}^*$ is determined through an inner maximization process that utilizes a non-convex loss function $\mathcal{L}_{adv}$.
We employ a perceptual metric, structural similarity index measure \citep[SSIM,][]{hore2010image}, as $\mathcal{L}_{adv}$ to explore more potential solutions and better handle complex perturbation patterns.
While $\boldsymbol{\delta}^*$ does not exactly represent the true adversarial perturbation, bounding $\norm{\boldsymbol{\delta}}<\eta$ can partially ensure that the misalignment between $\vect{y}$ and $\vect{x}_{adv}$ remains controlled, effectively ensuring that $\vect{y}$ does not simply collapse into the adversarial example $\vect{x}_{adv}$.

However, also since $\boldsymbol{\delta}^*$ does not represent the true perturbation, minimizing $||\vect{x}_d - (\hat{\vect{y}}+\boldsymbol{\delta}^*)||_2$ may not yield the desired clean reconstructed example. To address this limitation, we introduce a second loss term $||\text{P}_{d} (\vect{y}_{d-1}) - \hat{\vect{y}}||_2$.
Specifically, we utilize the reconstructed downsampled example $\vect{y}_{d-1}$ as an additional prior constraint to aid in approximating the $\vect{x}_{cln})$.
Building upon the observations in \Cref{fig:distribution}, we start from the resolution $D-l$ example $\vect{x}_{D-l}$, optimized by PuTT, and then perform upsampling to the higher resolution to produce a ``clean-leaning'' reference, which nudges $\vect{y}$ toward a less perturbed distribution.
Although the clean example $\vect{x}_{cln}$ cannot be obtained as a prior for the optimization, we devise a clever way to provide a surrogate prior and guide the optimization process.
The detailed algorithm of our adversarial optimization process is shown in \cref{alg:ours}.

% \textbf{Theorem 2.} \gl{To support loss, maybe a smaller bound between (x, y) compared with org loss. If we still have time.}

\section{Experiments}
\label{Experiments}
\renewcommand{\arraystretch}{0.8}

In this section, we conduct extensive experiments on CIFAR-10, CIFAR-100, and ImageNet across various attack settings to evaluate the performance of our method. The classification results demonstrate that the proposed method achieves state-of-the-art performance and exhibits strong generalization capabilities.
Specifically, our method achieved a 26.45\% improvement in average robust accuracy over AT across different norm threats, a 9.39\% improvement over AP across multiple attacks, and a 6.47\% improvement over AP across different datasets.
Next, we investigate the effectiveness of adversarial perturbation removal in denoising tasks using the existing tensor network decomposition methods. Among these, only our method successfully removes adversarial perturbations while preserving consistency between the reconstructed clean example and the reconstructed adversarial example.
These results collectively highlight the effectiveness and potential of our proposed method.

\subsection{Experimental setup}
\textbf{Datasets and model architectures} \quad We conduct extensive experiments on CIFAR-10, CIFAR-100 \citep{krizhevsky2009learning} and ImageNet \citep{deng2009imagenet} to empirically validate the effectiveness of the proposed methods against adversarial attacks. For classification tasks, we utilize the pre-trained ResNet \citep{he2016deep} and WideResNet \citep{zagoruyko2016wide} models. For denoising tasks, we employ Tensor Train \citep[TT,][]{oseledets2011tensor}, Tensor Ring \citep[TR,][]{zhao2016tensor}, quantized technique \citep{khoromskij2011d} and PuTT \citep{loeschcke2024coarse}.

\textbf{Adversarial attacks} \quad We evaluate our method against AutoAttack \citep{croce2020reliable}, a widely used benchmark that integrates both white-box and black-box attacks. Additionally, following the guidance of \citet{lee2023robust}, we utilize projected gradient descent \citep[PGD,][]{madry2018towards} with expectation over time \citep[EOT,][]{athalye2018synthesizing} for a more comprehensive evaluation.

% \textbf{Evaluation metrics:} We evaluate the performance of defense methods using multiple metrics: Standard accuracy and robust accuracy \citep{szegedy2013intriguing} on classification tasks. NRMSE, SSIM, PSNR metrics \citep{loeschcke2024coarse} and visualization results on denoising tasks.

Due to the high computational cost of evaluating methods with multiple attacks, following the guidance of \citet{nie2022diffusion}, we randomly select 512 images from the test set for robust evaluation. All experiments presented in the paper are conducted by NVIDIA RTX A5000 with 24GB GPU memory, CUDA v11.7 and cuDNN v8.5.0 in PyTorch v1.13.11 \citep{paszke2019pytorch}. More details in \Cref{app:settings}.

\subsection{Comparison with the state-of-the-art methods}
\label{Comparison results}

\begin{table}[t]
\centering
\caption{Standard and robust accuracy against AutoAttack $l_\infty$ threat ($\epsilon=8/255$) on CIFAR-10. ($^{\text{\textdagger}}$the methods use additional synthetic images. $^*$use robust classifer \citep{cui2024decoupled}.)}
\vskip 0.15in
\label{tab:cifar10:linf}
\begin{tabular}{@{\hspace{8pt}}c@{\hspace{18pt}}c@{\hspace{17pt}}c@{\hspace{17pt}}c@{\hspace{8pt}}}
\toprule
\multirow{2}{*}{Defense method} & Extra & Standard & Robust \\
& data & Acc. & Acc. \\
\midrule
\citet{zhang2020geometry} & \checkmark & 85.36  & 59.96  \\
\citet{gowal2020uncovering} & \checkmark & 89.48  & 62.70  \\
\citet{bai2023improving} & $\checkmark^{\text{\textdagger}}$ & 95.23  & 68.06 \\
\midrule
\citet{rebuffi2021fixing} & $\times^{\text{\textdagger}}$ & 87.33  & 61.72  \\
\citet{gowal2021improving} & $\times^{\text{\textdagger}}$ & 88.74  & 66.11 \\
\citet{wang2023better} & $\times^{\text{\textdagger}}$ & 93.25  & 70.69 \\
\citet{peng2023robust} & $\times^{\text{\textdagger}}$ & 93.27  & 71.07 \\
\citet{cui2024decoupled} & $\times^{\text{\textdagger}}$ & 92.16  & 67.73 \\
\midrule
\citet{nie2022diffusion}   & $\times$ & 89.02  & 70.64 \\
\citet{wang2022guided}  & $\times$ & 84.85  & 71.18 \\
\citet{anonymous2023classifier}  & $\times$ & 90.04  & 73.05 \\
\citet{lin2024adversarial} & $\times$ & 90.62  & 72.85 \\
Ours & $\times$ & 82.23  & 55.27  \\
Ours$^*$ & $\times$ & 91.99  & 72.85  \\
\bottomrule
\bottomrule
\end{tabular}
\vskip -0.1in
\end{table}

\begin{table}[t]
    \caption{Standard and robust accuracy against AutoAttack $l_2$ threat ($\epsilon=0.5$) on CIFAR-10. ($^{\text{\textdagger}}$the methods use additional synthetic images. $^*$use robust classifer \citep{cui2024decoupled}.)}
    \vskip 0.15in
    \label{tab:cifar10:l2}
    \begin{tabular}{@{\hspace{7pt}}c@{\hspace{16pt}}c@{\hspace{15pt}}c@{\hspace{15pt}}c@{\hspace{8pt}}}
    \toprule
    \multirow{2}{*}{Defense method} & Extra & Standard & Robust \\
    & data & Acc. & Acc. \\
    \midrule
    \citet{augustin2020adversarial} & \checkmark & 92.23  & 77.93  \\
    \citet{gowal2020uncovering} & \checkmark & 94.74  & 80.53  \\
    \midrule
    % \citet{wang2023better} & $\times^{\text{\textdagger}}$ & 95.16  & 83.68  \\
    \citet{rebuffi2021fixing} & $\times^{\text{\textdagger}}$ & 91.79  & 78.32  \\
    \midrule
    \citet{ding2019mma} & $\times$ & 88.02  & 67.77  \\
    \citet{nie2022diffusion} & $\times$ & 91.03 & 78.58  \\
    % \citet{anonymous2023classifier}  & $\times$ & 92.58  & 83.13  \\
    % \citet{bai2024diffusion} & $\times$ & 93.75 & 84.38 \\
    % \citet{lee2023robust}  & $\times$ & 90.16 & 86.48  \\
    Ours & $\times$ & 82.23  & 68.16  \\
    Ours$^*$ & $\times$ & 91.99  & 79.49  \\
    \bottomrule
    \bottomrule
    \end{tabular}%
    \vskip -0.1in
\end{table}

\begin{table}[htbp]
    \centering
    \caption{Standard and robust accuracy against AutoAttack $l_\infty$ threat ($\epsilon=8/255$) on CIFAR-100 using WideResNet-28-10 classifier. ($^{\text{\textdagger}}$the methods use additional synthetic images.)}
    \vskip 0.15in
    \label{tab:cifar100:linf}
    \begin{tabular}{c@{\hspace{12pt}}c@{\hspace{13pt}}c@{\hspace{13pt}}c}
    \toprule
    \multirow{2}{*}{Defense method} & Extra & Standard & Robust \\
    & data & Acc. & Acc. \\
    \midrule
    \citet{hendrycks2019using} & \checkmark & 59.23  & 28.42  \\
    \citet{debenedetti2023light} & \checkmark & 70.76 & 35.08 \\
    \midrule
    \citet{cui2024decoupled} & $\times^{\text{\textdagger}}$ & 73.85  & 39.18  \\
    \citet{wang2023better} & $\times^{\text{\textdagger}}$ & 75.22  & 42.67  \\
    \midrule
    \citet{pang2022robustness} & $\times$ & 63.66  & 31.08  \\
    \citet{jia2022adversarial} & $\times$ & 67.31 & 31.91 \\
    \citet{cui2024decoupled} & $\times$ & 65.93  & 32.52  \\
    Ours & $\times$ & 62.30  & 44.34  \\
    % Ours$^*$ & $\times$ & 71.48  & 44.53  \\
    \bottomrule
    \bottomrule
    \end{tabular}
    \vskip -0.1in
\end{table}

\begin{table}[t]
    \centering
    \caption{Standard and robust accuracy against AutoAttack $l_\infty$ threat ($\epsilon=4/255$) on ImageNet using ResNet-50 classifier.}
    \vskip 0.15in
    \label{tab:image:linf}
    \begin{tabular}{@{\hspace{6pt}}c@{\hspace{15pt}}c@{\hspace{15pt}}c@{\hspace{15pt}}c@{\hspace{8pt}}}
    \toprule
    \multirow{2}{*}{Defense method} & Extra & Standard & Robust \\
    & data & Acc. & Acc. \\
    \midrule
    \citet{engstrom2019robustness} & $\times$ & 62.56 & 31.06  \\
    \citet{wong2020fast} & $\times$ & 55.62 & 26.95  \\
    \citet{salman2020adversarially} & $\times$ & 64.02 & 37.89  \\
    \citet{bai2021transformers} & $\times$ & 67.38 & 35.51  \\
    \citet{nie2022diffusion} & $\times$ & 67.79 & 40.93  \\
    % \citet{bai2024diffusion} & $\times$ & 70.41 & 41.70  \\
    \citet{chen2024data} & $\times$ & 68.76 & 40.60 \\
    Ours & $\times$ & 65.43  & 42.77  \\
    % Ours$^*$ & $\times$ & -  & \textbf{-}  \\
    \bottomrule
    \bottomrule
    \end{tabular}
    \vskip -0.1in
\end{table}

In this section,
we evaluate our method for defending against AutoAttack $l_\infty$ and $l_2$ threats \citep{croce2020reliable,croce2021robustbench} and compare it with the state-of-the-art methods under all adversarial settings listed in RobustBench.
% \footnote{https://robustbench.github.io}.
\Cref{tab:cifar10:linf,tab:cifar10:l2,tab:cifar100:linf,tab:image:linf} present the performance of various defense methods against AutoAttack $l_\infty$ ($\epsilon=8/255$) and $l_2$ ($\epsilon=0.5$) threats on CIFAR-10, CIFAR-100 and ImageNet datasets.
Overall, the highest robust accuracy achievable by our method is generally on par with the current state-of-the-art methods without using extra data (the dataset introduced by \citet{carmon2019unlabeled}). Specifically, compared to the second-best method, our method improves the robust accuracy by 1.67\% on CIFAR-100, by 1.84\% on ImageNet, and the average robust accuracy by 0.36\% on CIFAR-10.

However, due to the overfitting of WideResNet-28-10 trained on the limited data available in CIFAR-10, we observe that the results of $\text{Ours}$ struggle to reach state-of-the-art performance, consistent with findings from other AT methods. To address this issue, these methods incorporate additional synthetic data to train a robust classifier. Following this, we conduct experiments using the robust classifier trained by \citet{cui2024decoupled}, which utilizes an additional 20M synthetic images in training. This leads to a significant improvement in the robust accuracy observed in $\text{Ours}^*$. 
Moreover, compared to the used robust classifier, our method futher improves the robust accuracy by 5.12\%.
These results are consistent across multiple datasets and norm threats, confirming the effectiveness of our method and its potential for defending against adversarial attacks.

\subsection{Generalization comparison across various scenarios}
\label{Comparison across multiple cases}
As previously mentioned, the existing defense methods are often criticized for their lack of generalization across different norm threats, attacks, and datasets. In this section, we evaluate the performance of our method under various settings to demonstrate its generalization capability.

\begin{figure*}[t]
\vskip -0.1in
  \begin{minipage}[ht]{0.65\linewidth}
  \centering
    \captionof{table}{Standard accuracy (SA) and robust accuracy (RA) against AutoAttack $l_\infty$ ($\epsilon=8/255$) threat on CIFAR-10 and CIFAR-100 with WideResNet-28-10 classifier. The pre-trained generative model used in AP is trained on CIFAR-10.}
    \vskip 0.15in
    \label{tab:datasets}
    \begin{tabular}{ccccccc}
    \toprule
    \multirow{2}{*}{Defense method} & \multicolumn{2}{c}{CIFAR-10} & \multicolumn{2}{c}{CIFAR-100} & \multicolumn{2}{c}{Avg.} \\
    \cmidrule(lr){2-3}
    \cmidrule(lr){4-5}
    \cmidrule(lr){6-7}
     & SA & RA & SA & RA & SA & RA\\
    \midrule
    Standard Training & 94.78  & 0.00  & 81.86 & 0.00 & 88.32 & 0.00 \\
    \midrule
    AT \citep{cui2024decoupled} & 92.16  & 67.73 & 73.85 & 39.18 & 83.01 & 53.46 \\
    AP \citep{nie2022diffusion} & 89.02  & 70.64 & 38.09 & 33.79 & 63.56 & 52.22\\
    Ours$^*$  & 91.99 & 72.85 & 71.48 & 44.53 & 81.74 & 58.69 \\
    \bottomrule
    \bottomrule
    \end{tabular}
    \vskip -0.1in
  \end{minipage}\quad\quad
  \begin{minipage}[ht]{0.3\linewidth}
  \vskip 0.2in
    \centering
    \includegraphics[width = \linewidth]{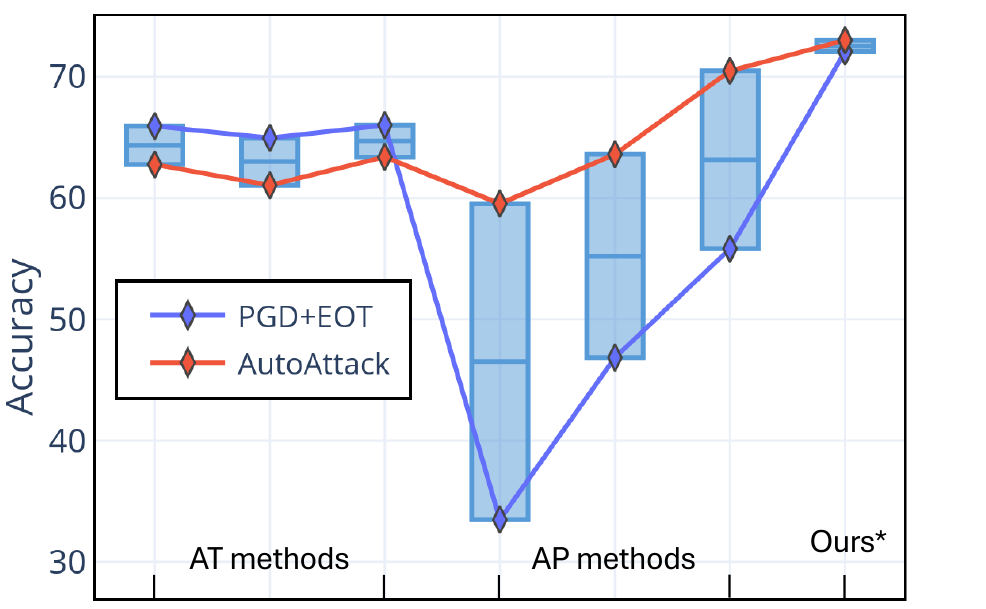}
    \captionof{figure}{Comparison of robust accuracy against PGD+EOT and AutoAttack.}
    \label{fig:attacks}
    \vskip -0.2in
  \end{minipage}
  \vskip -0.1in
\end{figure*}

\textbf{Results analysis on different norm-threats:}
\Cref{tab:norms} shows that AT methods \citep{laidlaw2021perceptual,dolatabadi2022} are limited in defending against unseen attacks and can only effectively against the specific attacks they are trained on. The intuitive idea is to apply AT across all norm threats or develop more general constraints to obtain a robust model. However, training such a model is challenging due to the inherent differences among various attacks. In contrast, AP methods \citep{nie2022diffusion,lin2024adversarial} exhibit strong generalization, effectively defending against unseen attacks. The results demonstrate that our method also possesses strong generalization capabilities against unseen attacks, achieving performance close to the state-of-the-art AP methods while significantly outperforming the existing AT methods. Specifically, compared to the best AT method, our method improves average robust accuracy by 26.45\%.

\begin{table}[tb]
\vskip -0.08in
\caption{Standard accuracy and robust accuracy against AutoAttack $l_\infty$ ($\epsilon=8/255$) and AutoAttack $l_2$ ($\epsilon=1.0$) threats on CIFAR-10 with standard ResNet-50 classifier.}
\vskip 0.15in
\label{tab:norms}
\begin{center}
\begin{tabular}{@{\hspace{3pt}}c@{\hspace{1pt}}c@{\hspace{12pt}}c@{\hspace{18pt}}cc@{\hspace{3pt}}}
    \toprule
    \multirow{2}{*}{Type} & \multirow{2}{*}{Defense method} & \multirow{2}{*}{SA} & \multicolumn{2}{c}{Robust Acc.} \\
    \cmidrule(){4-5}
    & &  & AA $l_\infty$ & AA $l_2$ \\
    \midrule
    & Standard Training & 94.8  & 0.0   & 0.0   \\
    \midrule
    \multirow{4}{*}{AT} & Training with $l_\infty$ & 86.8  & 49.0  & 19.2  \\
    & Training with $l_2$ & 85.0  & 39.5  & 47.8  \\
    & \citet{laidlaw2021perceptual} & 82.4  & 30.2  & 34.9  \\
    & \citet{dolatabadi2022} & 83.2  & 40.0  & 33.9  \\
    \midrule
    \multirow{2}{*}{AP} & \citet{nie2022diffusion} & 88.2  & 70.0  & 70.9  \\
    & \citet{lin2024adversarial} & 89.1  & 71.2 & 73.4  \\
    \midrule
    & Ours  & 88.3 & 73.2 & 67.0 \\
    \bottomrule
    \bottomrule
\end{tabular}
\end{center}
\vskip -0.2in
\end{table}

\begin{figure*}[t]
\vskip 0.05in
    \centering
    \includegraphics[width=\linewidth]{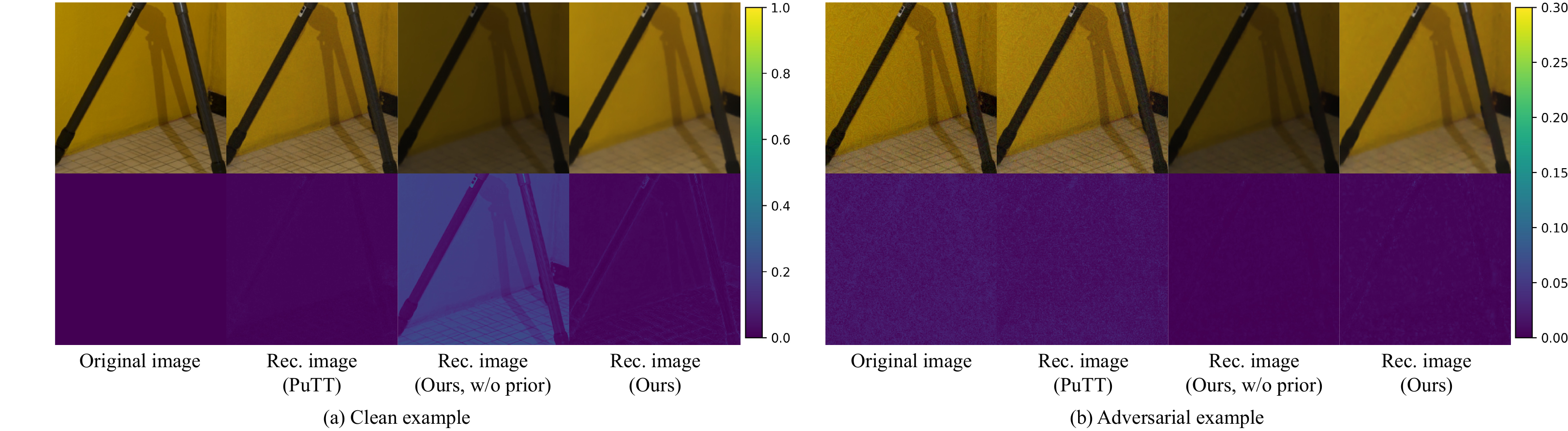}
    \vskip -0.1in
    \caption{Visual comparison of the denoising task. Top: the original image and corresponding reconstructed image for (a) the clean example and (b) the adversarial example, using PuTT and our proposed method. Bottom: the error maps are created (a) between the rec. clean example and the original clean example, as well as (b) between the rec. adversarial example and the rec. clean example.}
    \label{fig:visual}
    \vskip -0.13in
\end{figure*}

\begin{table*}[t]
    \caption{Performance comparison of various methods on the denoising task. We evaluate the accuracy, NRMSE, SSIM and PSNR metrics using clean examples and adversarial examples on CIFAR-10. Additionally, we compare the differences between rec. AEs and rec. CEs.}
    \vskip 0.15in
    \label{tab:reconstruction}
    \begin{center}
    \begin{tabular}{ccccccccccccc}
    \toprule
    Defense & \multicolumn{4}{c}{CLN: CEs $\text{ }\&\text{ }$ rec. CEs} & \multicolumn{4}{c}{ADV: AEs $\text{ }\&\text{ }$ rec. AEs} & \multicolumn{3}{c}{REC: rec. CEs $\text{ }\&\text{ }$ rec. AEs} \\
    \cmidrule(lr){2-5}
    \cmidrule(lr){6-9}
    \cmidrule(lr){10-12}
    method & Acc. & NRMSE & SSIM & PSNR & Acc. & NRMSE & SSIM & PSNR & NRMSE & SSIM & PSNR\\
    % method & SA & NRMSE$\downarrow$ & SSIM$\uparrow$ & PSNR$\uparrow$ & RA & NRMSE$\uparrow$ & SSIM$\downarrow$ & PSNR$\downarrow$ & NRMSE$\downarrow$ & SSIM$\uparrow$ & PSNR$\uparrow$ \\
    \midrule
    Standard & 94.73 & - & - & - & 0.00 & - & - & - & - & - & - \\
    \midrule
    TT & 87.30 & 0.0507 & 0.9526 & 31.14 & 36.13 & 0.065 & 0.8977 & 28.99 & 0.0267 & 0.9790 & 39.10 \\
    TR & \textbf{94.34} & \textbf{0.0171} & \textbf{0.9938} & \textbf{40.58} & 0.98 & 0.0464 & 0.9210 & 31.91 & 0.0322 & 0.9598 & 35.51 \\
    QTT & 84.57 & 0.0613 & 0.9253 & 29.49 & 51.56 & 0.0724 & 0.8808 & 28.06 & 0.0233 & 0.9855 & 39.88 \\
    QTR & 83.40 & 0.0613 & 0.9254 & 29.49 & 49.41 & 0.0724 & 0.8785 & 28.06 & 0.0231 & 0.9853 & 39.96 \\
    PuTT & 80.86 & 0.0626 & 0.9261 & 29.32 & 44.14 & 0.0742 & 0.8787 & 27.84 & 0.0311 & 0.9770 & 38.03 \\
    \midrule
    Ours & 82.23 & 0.0644 & 0.9203 & 29.06 & \textbf{55.27} & \textbf{0.0748} & \textbf{0.8707} & \textbf{27.77} & \textbf{0.0218} & \textbf{0.9863} & \textbf{40.37}  \\
    \bottomrule
    \bottomrule
    \end{tabular}
    \end{center}
\vskip -0.1in
\end{table*}

\textbf{Results analysis on multiple attacks:} \cref{fig:attacks} shows the comparison of robust accuracy against PGD+EOT and AutoAttack with $l_\infty$ ($\epsilon=8/255$) threat on CIFAR-10 with WideResNet-28-10. When facing different attacks within the same threat, AT methods \citep{gowal2020uncovering,gowal2021improving,pang2022robustness} exhibit better generalization than AP methods \citep{yoon2021adversarial,nie2022diffusion,lee2023robust}. Typically, robustness evaluation is based on the worst-case results of the robust accuracy. Under this criterion, our method outperforms all AT and AP methods. Furthermore, compared to the state-of-the-art AP method on both attacks, our method improves the average robust accuracy by 9.39\%.

\textbf{Results analysis on different datasets:} \cref{tab:datasets} shows the generalization of the methods across different datasets. As previously mentioned, the existing AP methods typically rely on the specific datasets. When a pre-trained generative model trained on CIFAR-10 is applied to adversarial robustness evaluation on CIFAR-100, both standard accuracy and robust accuracy on CIFAR-100 drop significantly. This occurs because the pre-trained generative model can only generate the data it has learned. Although the input examples originate from CIFAR-100, the generative model attempts to output one of the ten classes from CIFAR-10, severely distorting the semantic information of the input examples and leading to low classification accuracy. In contrast, TN-based AP methods rely solely on the input examples rather than prior information learned from large datasets, allowing them generalize effectively across different datasets. The results demonstrate that our method exhibits strong generalization across different datasets, achieving comparable robust performance on CIFAR-100 as on CIFAR-10. Specifically, compared to the AP method \citep{nie2022diffusion}, our method improves the average robust accuracy by 6.47\%.

\subsection{Denoising tasks}
\label{Denoising tasks}

In this section, we evaluate the effectiveness of our method on non-classification tasks through visual comparisons and various quantitative metrics.

\textbf{Visualization results analysis:} \Cref{fig:visual} shows the visual comparison of the denoising task on ImageNet. The top row in (a) displays the input clean example (CE), and its corresponding reconstructed clean examples (rec. CE) generated by PuTT and our proposed method, while (b) displays the reconstructed adversarial examples (rec. AE) for the input adversarial example (AE).
Additionally, we create error maps to highlight differences, as shown at the bottom of \Cref{fig:visual}: (a) between the rec. CEs and the input CEs, and (b) between the rec. AEs and the rec. CEs. The results indicate that while our method does not match PuTT in reconstructing CEs, it significantly outperforms PuTT in removing adversarial perturbations from AEs.

Specifically, when processing CEs, the reconstructed examples generated by PuTT are almost identical to the original ones, whereas our method is slightly less effective in restoring some details. However, when processing AEs, the reconstructed examples from PuTT remain consistent with the original ones, leading to the preservation of adversarial perturbations, as highlighted in \Cref{fig:visual}b. In contrast, our method better removes those perturbations, ensuring that the rec. AEs and the rec. CEs retain similar information. Moreover, we evaluate the necessity of the second term in \Cref{eq:ours}, which serves as a surrogate prior constraint to optimize the reconstructed examples toward the clean data distribution. As observed, removing this constraint eliminates prior information from the optimization process, increasing the likelihood of significant deviation in the wrong direction.

\textbf{Quantitative results analysis:} \Cref{tab:reconstruction} shows the quantitative results of the denoising task for AEs and CEs of CIFAR-10. We compare our method with existing tensor network decomposition methods, including TT, TR, QTT, QTR, and PuTT.
While our method does not achieve the best denoising performance on clean examples, it still maintains classification performance well, achieving 82.23\% standard accuracy with the vanilla WideResNet-28-10 classifier. More importantly, our method outperforms others in the next two columns. Specifically, when processing AEs, our method yields the highest NRMSE and the lowest SSIM and PSNR, achieving the highest robust accuracy.
This outcome is expected, as our goal is to ensure that the rec. AEs differ from the original AEs (i.e., lower SSIM and PSNR, and higher NRMSE in the ``ADV'' column) while rec. AEs closely resembling the rec. CEs (i.e., higher SSIM and PSNR, and lower NRMSE in the ``REC'' column).
These results align well with the visual observations in \Cref{fig:visual} and consistently demonstrate the effectiveness of our method highlighting its potential in adversarial scenarios.

\textbf{Limitations} \quad One limitation of our method is that, despite being a training-free technique, TN-based AP method requires additional optimization time during inference.
This overhead stems from the inherent limitations of iterative optimization processes and impacts their practicality in real-world applications.
Therefore, we leave the exploration of integrating our TN-based AP technique with more advanced and efficient optimization strategies for future research.

\section{Conclusion}
\label{Conclusion}
In this paper, we propose a novel model-free adversarial purification method based on a specially designed tensor network decomposition algorithm. We conduct extensive experiments on CIFAR-10, CIFAR-100, and ImageNet, demonstrating that our method achieves state-of-the-art performance in defending against adversarial attacks while exhibiting strong generalization across diverse adversarial scenarios.
Despite these significant improvements, our method features an additional optimization cost during inference.
However, further exploration of TN-based AP method remains an exciting research direction for developing a plug-and-play and effective adversarial purification technique.

% \section*{Acknowledgements}

\clearpage
\section*{Impact Statement}
This paper presents work whose goal is to advance the field of
Machine Learning. There are many potential societal consequences
of our work, none which we feel must be specifically highlighted here.

% In the unusual situation where you want a paper to appear in the references without citing it in the main text, use \nocite

\bibliography{icml2025}

\begin{thebibliography}{68}
\providecommand{\natexlab}[1]{#1}
\providecommand{\url}[1]{\texttt{#1}}
\expandafter\ifx\csname urlstyle\endcsname\relax
  \providecommand{\doi}[1]{doi: #1}\else
  \providecommand{\doi}{doi: \begingroup \urlstyle{rm}\Url}\fi

\bibitem[Allen-Zhu \& Li(2022)Allen-Zhu and Li]{allen2022feature}
Allen-Zhu, Z. and Li, Y.
\newblock Feature purification: How adversarial training performs robust deep learning.
\newblock In \emph{2021 IEEE 62nd Annual Symposium on Foundations of Computer Science (FOCS)}, pp.\  977--988. IEEE, 2022.

\bibitem[Athalye et~al.(2018)Athalye, Engstrom, Ilyas, and Kwok]{athalye2018synthesizing}
Athalye, A., Engstrom, L., Ilyas, A., and Kwok, K.
\newblock Synthesizing robust adversarial examples.
\newblock In \emph{International conference on machine learning}, pp.\  284--293. PMLR, 2018.

\bibitem[Augustin et~al.(2020)Augustin, Meinke, and Hein]{augustin2020adversarial}
Augustin, M., Meinke, A., and Hein, M.
\newblock Adversarial robustness on in-and out-distribution improves explainability.
\newblock In \emph{European Conference on Computer Vision}, pp.\  228--245. Springer, 2020.

\bibitem[Bai et~al.(2024)Bai, Huang, Li, Wang, Gao, Caiafa, and Zhao]{bai2024diffusion}
Bai, M., Huang, W., Li, T., Wang, A., Gao, J., Caiafa, C.~F., and Zhao, Q.
\newblock Diffusion models demand contrastive guidance for adversarial purification to advance.
\newblock In \emph{Forty-first International Conference on Machine Learning}, 2024.

\bibitem[Bai et~al.(2021)Bai, Mei, Yuille, and Xie]{bai2021transformers}
Bai, Y., Mei, J., Yuille, A.~L., and Xie, C.
\newblock Are transformers more robust than cnns?
\newblock \emph{Advances in neural information processing systems}, 34:\penalty0 26831--26843, 2021.

\bibitem[Bai et~al.(2023)Bai, Anderson, Kim, and Sojoudi]{bai2023improving}
Bai, Y., Anderson, B.~G., Kim, A., and Sojoudi, S.
\newblock Improving the accuracy-robustness trade-off of classifiers via adaptive smoothing.
\newblock \emph{arXiv preprint arXiv:2301.12554}, 2023.

\bibitem[Bhattarai et~al.(2023)Bhattarai, Kaymak, Barron, Nebgen, Rasmussen, and Alexandrov]{bhattarai2023robust}
Bhattarai, M., Kaymak, M.~C., Barron, R., Nebgen, B., Rasmussen, K., and Alexandrov, B.~S.
\newblock Robust adversarial defense by tensor factorization.
\newblock In \emph{2023 International Conference on Machine Learning and Applications (ICMLA)}, pp.\  308--315. IEEE, 2023.

\bibitem[Botchkarev(2018)]{botchkarev2018performance}
Botchkarev, A.
\newblock Performance metrics (error measures) in machine learning regression, forecasting and prognostics: Properties and typology.
\newblock \emph{arXiv preprint arXiv:1809.03006}, 2018.

\bibitem[Carmon et~al.(2019)Carmon, Raghunathan, Schmidt, Duchi, and Liang]{carmon2019unlabeled}
Carmon, Y., Raghunathan, A., Schmidt, L., Duchi, J.~C., and Liang, P.~S.
\newblock Unlabeled data improves adversarial robustness.
\newblock \emph{Advances in neural information processing systems}, 32, 2019.

\bibitem[Chen \& Lee(2024)Chen and Lee]{chen2024data}
Chen, E.-C. and Lee, C.-R.
\newblock Data filtering for efficient adversarial training.
\newblock \emph{Pattern Recognition}, 151:\penalty0 110394, 2024.

\bibitem[Cichocki et~al.(2015)Cichocki, Mandic, De~Lathauwer, Zhou, Zhao, Caiafa, and Phan]{cichocki2015tensor}
Cichocki, A., Mandic, D., De~Lathauwer, L., Zhou, G., Zhao, Q., Caiafa, C., and Phan, H.~A.
\newblock Tensor decompositions for signal processing applications: From two-way to multiway component analysis.
\newblock \emph{IEEE signal processing magazine}, 32\penalty0 (2):\penalty0 145--163, 2015.

\bibitem[Croce \& Hein(2020)Croce and Hein]{croce2020reliable}
Croce, F. and Hein, M.
\newblock Reliable evaluation of adversarial robustness with an ensemble of diverse parameter-free attacks.
\newblock In \emph{International conference on machine learning}, pp.\  2206--2216. PMLR, 2020.

\bibitem[Croce et~al.(2021)Croce, Andriushchenko, Sehwag, Debenedetti, Flammarion, Chiang, Mittal, and Hein]{croce2021robustbench}
Croce, F., Andriushchenko, M., Sehwag, V., Debenedetti, E., Flammarion, N., Chiang, M., Mittal, P., and Hein, M.
\newblock Robustbench: a standardized adversarial robustness benchmark.
\newblock In \emph{Thirty-fifth Conference on Neural Information Processing Systems Datasets and Benchmarks Track (Round 2)}, 2021.

\bibitem[Cui et~al.(2024)Cui, Tian, Zhong, QI, Yu, and Zhang]{cui2024decoupled}
Cui, J., Tian, Z., Zhong, Z., QI, X., Yu, B., and Zhang, H.
\newblock Decoupled kullback-leibler divergence loss.
\newblock In \emph{The Thirty-eighth Annual Conference on Neural Information Processing Systems}, 2024.

\bibitem[Dai et~al.(2020)Dai, Feng, Wu, Chen, Lu, Jiang, and Xia]{dipdefend2020}
Dai, T., Feng, Y., Wu, D., Chen, B., Lu, J., Jiang, Y., and Xia, S.-T.
\newblock Dipdefend: Deep image prior driven defense against adversarial examples.
\newblock In \emph{Proceedings of the 28th ACM International Conference on Multimedia}, MM '20, pp.\  1404–1412, New York, NY, USA, 2020. Association for Computing Machinery.
\newblock ISBN 9781450379885.
\newblock \doi{10.1145/3394171.3413898}.
\newblock URL \url{https://doi.org/10.1145/3394171.3413898}.

\bibitem[Dai et~al.(2022)Dai, Feng, Chen, Lu, and Xia]{dai2022deep}
Dai, T., Feng, Y., Chen, B., Lu, J., and Xia, S.-T.
\newblock Deep image prior based defense against adversarial examples.
\newblock \emph{Pattern Recognition}, 122:\penalty0 108249, 2022.

\bibitem[Debenedetti et~al.(2023)Debenedetti, Sehwag, and Mittal]{debenedetti2023light}
Debenedetti, E., Sehwag, V., and Mittal, P.
\newblock A light recipe to train robust vision transformers.
\newblock In \emph{2023 IEEE Conference on Secure and Trustworthy Machine Learning (SaTML)}, pp.\  225--253. IEEE, 2023.

\bibitem[Deng et~al.(2009)Deng, Dong, Socher, Li, Li, and Fei-Fei]{deng2009imagenet}
Deng, J., Dong, W., Socher, R., Li, L.-J., Li, K., and Fei-Fei, L.
\newblock Imagenet: A large-scale hierarchical image database.
\newblock In \emph{2009 IEEE conference on computer vision and pattern recognition}, pp.\  248--255. Ieee, 2009.

\bibitem[Ding et~al.(2019)Ding, Sharma, Lui, and Huang]{ding2019mma}
Ding, G.~W., Sharma, Y., Lui, K. Y.~C., and Huang, R.
\newblock Mma training: Direct input space margin maximization through adversarial training.
\newblock In \emph{International Conference on Learning Representations}, 2019.

\bibitem[Dolatabadi et~al.(2022)Dolatabadi, Erfani, and Leckie]{dolatabadi2022}
Dolatabadi, H.~M., Erfani, S., and Leckie, C.
\newblock l-inf robustness and beyond: Unleashing efficient adversarial training.
\newblock In \emph{European Conference on Computer Vision}, pp.\  467--483. Springer, 2022.

\bibitem[Engstrom et~al.(2019)Engstrom, Ilyas, Salman, Santurkar, and Tsipras]{engstrom2019robustness}
Engstrom, L., Ilyas, A., Salman, H., Santurkar, S., and Tsipras, D.
\newblock Robustness (python library), 2019.
\newblock \emph{URL https://github. com/MadryLab/robustness}, 4\penalty0 (4):\penalty0 4--3, 2019.

\bibitem[Entezari \& Papalexakis(2022)Entezari and Papalexakis]{entezari2022tensorshield}
Entezari, N. and Papalexakis, E.~E.
\newblock Tensorshield: Tensor-based defense against adversarial attacks on images.
\newblock In \emph{MILCOM 2022-2022 IEEE Military Communications Conference (MILCOM)}, pp.\  999--1004. IEEE, 2022.

\bibitem[Goodfellow et~al.(2015)Goodfellow, Shlens, and Szegedy]{goodfellow2014explaining}
Goodfellow, I.~J., Shlens, J., and Szegedy, C.
\newblock Explaining and harnessing adversarial examples.
\newblock \emph{International Conference on Learning Representations}, 2015.

\bibitem[Gowal et~al.(2020)Gowal, Qin, Uesato, Mann, and Kohli]{gowal2020uncovering}
Gowal, S., Qin, C., Uesato, J., Mann, T., and Kohli, P.
\newblock Uncovering the limits of adversarial training against norm-bounded adversarial examples.
\newblock \emph{arXiv preprint arXiv:2010.03593}, 2020.

\bibitem[Gowal et~al.(2021)Gowal, Rebuffi, Wiles, Stimberg, Calian, and Mann]{gowal2021improving}
Gowal, S., Rebuffi, S.-A., Wiles, O., Stimberg, F., Calian, D.~A., and Mann, T.~A.
\newblock Improving robustness using generated data.
\newblock \emph{Advances in Neural Information Processing Systems}, 34:\penalty0 4218--4233, 2021.

\bibitem[Grzenda \& Zieba(2008)Grzenda and Zieba]{grzenda2008conditional}
Grzenda, W. and Zieba, W.
\newblock Conditional central limit theorem.
\newblock In \emph{Int. Math. Forum}, volume~3, pp.\  1521--1528, 2008.

\bibitem[He et~al.(2016)He, Zhang, Ren, and Sun]{he2016deep}
He, K., Zhang, X., Ren, S., and Sun, J.
\newblock Deep residual learning for image recognition.
\newblock In \emph{Proceedings of the IEEE conference on computer vision and pattern recognition}, pp.\  770--778, 2016.

\bibitem[He et~al.(2022)He, Chen, Xie, Li, Doll{\'a}r, and Girshick]{he2022masked}
He, K., Chen, X., Xie, S., Li, Y., Doll{\'a}r, P., and Girshick, R.
\newblock Masked autoencoders are scalable vision learners.
\newblock In \emph{Proceedings of the IEEE/CVF conference on computer vision and pattern recognition}, pp.\  16000--16009, 2022.

\bibitem[Hendrycks et~al.(2019)Hendrycks, Lee, and Mazeika]{hendrycks2019using}
Hendrycks, D., Lee, K., and Mazeika, M.
\newblock Using pre-training can improve model robustness and uncertainty.
\newblock In \emph{International conference on machine learning}, pp.\  2712--2721. PMLR, 2019.

\bibitem[Hore \& Ziou(2010)Hore and Ziou]{hore2010image}
Hore, A. and Ziou, D.
\newblock Image quality metrics: Psnr vs. ssim.
\newblock In \emph{2010 20th international conference on pattern recognition}, pp.\  2366--2369. IEEE, 2010.

\bibitem[Hubig et~al.(2017)Hubig, McCulloch, and Schollw{\"o}ck]{hubig2017generic}
Hubig, C., McCulloch, I., and Schollw{\"o}ck, U.
\newblock Generic construction of efficient matrix product operators.
\newblock \emph{Physical Review B}, 95\penalty0 (3):\penalty0 035129, 2017.

\bibitem[Ilyas et~al.(2019)Ilyas, Santurkar, Tsipras, Engstrom, Tran, and Madry]{ilyas2019adversarial}
Ilyas, A., Santurkar, S., Tsipras, D., Engstrom, L., Tran, B., and Madry, A.
\newblock Adversarial examples are not bugs, they are features.
\newblock \emph{Advances in neural information processing systems}, 32, 2019.

\bibitem[Jia et~al.(2022)Jia, Zhang, Wu, Ma, Wang, and Cao]{jia2022adversarial}
Jia, X., Zhang, Y., Wu, B., Ma, K., Wang, J., and Cao, X.
\newblock Las-at: adversarial training with learnable attack strategy.
\newblock In \emph{Proceedings of the IEEE/CVF Conference on Computer Vision and Pattern Recognition}, pp.\  13398--13408, 2022.

\bibitem[Khoromskij(2011)]{khoromskij2011d}
Khoromskij, B.~N.
\newblock O (d log n)-quantics approximation of n-d tensors in high-dimensional numerical modeling.
\newblock \emph{Constructive Approximation}, 34:\penalty0 257--280, 2011.

\bibitem[Kolda \& Bader(2009)Kolda and Bader]{kolda2009tensor}
Kolda, T.~G. and Bader, B.~W.
\newblock Tensor decompositions and applications.
\newblock \emph{SIAM review}, 51\penalty0 (3):\penalty0 455--500, 2009.

\bibitem[Krizhevsky et~al.(2009)Krizhevsky, Hinton, et~al.]{krizhevsky2009learning}
Krizhevsky, A., Hinton, G., et~al.
\newblock Learning multiple layers of features from tiny images.
\newblock \emph{Technical Report}, 2009.

\bibitem[Laidlaw et~al.(2021)Laidlaw, Singla, and Feizi]{laidlaw2021perceptual}
Laidlaw, C., Singla, S., and Feizi, S.
\newblock Perceptual adversarial robustness: Defense against unseen threat models.
\newblock In \emph{International Conference on Learning Representations (ICLR)}, 2021.

\bibitem[Lee \& Kim(2023)Lee and Kim]{lee2023robust}
Lee, M. and Kim, D.
\newblock Robust evaluation of diffusion-based adversarial purification.
\newblock In \emph{Proceedings of the IEEE/CVF International Conference on Computer Vision (ICCV)}, pp.\  134--144, October 2023.

\bibitem[Lin et~al.(2024)Lin, Li, Zhang, Tanaka, and Zhao]{lin2024adversarial}
Lin, G., Li, C., Zhang, J., Tanaka, T., and Zhao, Q.
\newblock Adversarial training on purification (atop): Advancing both robustness and generalization.
\newblock \emph{arXiv preprint arXiv:2401.16352}, 2024.

\bibitem[Loeschcke et~al.(2024)Loeschcke, Wang, Leth-Espensen, Belongie, Kastoryano, and Benaim]{loeschcke2024coarse}
Loeschcke, S.~B., Wang, D., Leth-Espensen, C.~M., Belongie, S., Kastoryano, M., and Benaim, S.
\newblock Coarse-to-fine tensor trains for compact visual representations.
\newblock In \emph{Forty-first International Conference on Machine Learning}, 2024.

\bibitem[Lubasch et~al.(2018)Lubasch, Moinier, and Jaksch]{lubasch2018multigrid}
Lubasch, M., Moinier, P., and Jaksch, D.
\newblock Multigrid renormalization.
\newblock \emph{Journal of Computational Physics}, 372:\penalty0 587--602, 2018.

\bibitem[Lyu et~al.(2023)Lyu, Wu, Yin, and Luo]{lyu2023maedefense}
Lyu, W., Wu, M., Yin, Z., and Luo, B.
\newblock Maedefense: An effective masked autoencoder defense against adversarial attacks.
\newblock In \emph{2023 Asia Pacific Signal and Information Processing Association Annual Summit and Conference (APSIPA ASC)}, pp.\  1915--1922. IEEE, 2023.

\bibitem[Madry et~al.(2018)Madry, Makelov, Schmidt, Tsipras, and Vladu]{madry2018towards}
Madry, A., Makelov, A., Schmidt, L., Tsipras, D., and Vladu, A.
\newblock Towards deep learning models resistant to adversarial attacks.
\newblock In \emph{International Conference on Learning Representations}, 2018.

\bibitem[McCulloch(2008)]{mcculloch2008infinite}
McCulloch, I.~P.
\newblock Infinite size density matrix renormalization group, revisited.
\newblock \emph{arXiv preprint arXiv:0804.2509}, 2008.

\bibitem[Nie et~al.(2022)Nie, Guo, Huang, Xiao, Vahdat, and Anandkumar]{nie2022diffusion}
Nie, W., Guo, B., Huang, Y., Xiao, C., Vahdat, A., and Anandkumar, A.
\newblock Diffusion models for adversarial purification.
\newblock \emph{International Conference on Machine Learning}, 2022.

\bibitem[Oseledets(2011)]{oseledets2011tensor}
Oseledets, I.~V.
\newblock Tensor-train decomposition.
\newblock \emph{SIAM Journal on Scientific Computing}, 33\penalty0 (5):\penalty0 2295--2317, 2011.

\bibitem[Pang et~al.(2022)Pang, Lin, Yang, Zhu, and Yan]{pang2022robustness}
Pang, T., Lin, M., Yang, X., Zhu, J., and Yan, S.
\newblock Robustness and accuracy could be reconcilable by (proper) definition.
\newblock In \emph{International Conference on Machine Learning}, pp.\  17258--17277. PMLR, 2022.

\bibitem[Paszke et~al.(2019)Paszke, Gross, Massa, Lerer, Bradbury, Chanan, Killeen, Lin, Gimelshein, Antiga, et~al.]{paszke2019pytorch}
Paszke, A., Gross, S., Massa, F., Lerer, A., Bradbury, J., Chanan, G., Killeen, T., Lin, Z., Gimelshein, N., Antiga, L., et~al.
\newblock Pytorch: An imperative style, high-performance deep learning library.
\newblock \emph{Advances in neural information processing systems}, 32, 2019.

\bibitem[Peng et~al.(2023)Peng, Xu, Cornelius, Hull, Li, Duggal, Phute, Martin, and Chau]{peng2023robust}
Peng, S., Xu, W., Cornelius, C., Hull, M., Li, K., Duggal, R., Phute, M., Martin, J., and Chau, D.~H.
\newblock Robust principles: Architectural design principles for adversarially robust cnns.
\newblock \emph{British Machine Vision Conference (BMVC)}, 2023.

\bibitem[Phan et~al.(2020)Phan, Cichocki, Uschmajew, Tichavsk{\`y}, Luta, and Mandic]{phan2020tensor}
Phan, A.-H., Cichocki, A., Uschmajew, A., Tichavsk{\`y}, P., Luta, G., and Mandic, D.~P.
\newblock Tensor networks for latent variable analysis: Novel algorithms for tensor train approximation.
\newblock \emph{IEEE transactions on neural networks and learning systems}, 31\penalty0 (11):\penalty0 4622--4636, 2020.

\bibitem[Phan et~al.(2023)Phan, Yin, Sui, Yuan, and Zonouz]{phan2023cstar}
Phan, H., Yin, M., Sui, Y., Yuan, B., and Zonouz, S.
\newblock Cstar: towards compact and structured deep neural networks with adversarial robustness.
\newblock In \emph{Proceedings of the AAAI Conference on Artificial Intelligence}, volume~37, pp.\  2065--2073, 2023.

\bibitem[Rebuffi et~al.(2021)Rebuffi, Gowal, Calian, Stimberg, Wiles, and Mann]{rebuffi2021fixing}
Rebuffi, S.-A., Gowal, S., Calian, D.~A., Stimberg, F., Wiles, O., and Mann, T.
\newblock Fixing data augmentation to improve adversarial robustness.
\newblock \emph{arXiv preprint arXiv:2103.01946}, 2021.

\bibitem[Rudkiewicz et~al.(2024)Rudkiewicz, Ouerfelli, Finotello, Chaouai, and Tamaazousti]{rudkiewicz2024robustness}
Rudkiewicz, T., Ouerfelli, M., Finotello, R., Chaouai, Z., and Tamaazousti, M.
\newblock Robustness of tensor decomposition-based neural network compression.
\newblock In \emph{2024 IEEE International Conference on Image Processing (ICIP)}, pp.\  221--227. IEEE, 2024.

\bibitem[Salman et~al.(2020)Salman, Ilyas, Engstrom, Kapoor, and Madry]{salman2020adversarially}
Salman, H., Ilyas, A., Engstrom, L., Kapoor, A., and Madry, A.
\newblock Do adversarially robust imagenet models transfer better?
\newblock \emph{Advances in Neural Information Processing Systems}, 33:\penalty0 3533--3545, 2020.

\bibitem[Shi et~al.(2021)Shi, Holtz, and Mishne]{shi2021online}
Shi, C., Holtz, C., and Mishne, G.
\newblock Online adversarial purification based on self-supervision.
\newblock \emph{International Conference on Learning Representations}, 2021.

\bibitem[Song et~al.(2024)Song, Choi, and Han]{song2024training}
Song, M., Choi, J., and Han, B.
\newblock A training-free defense framework for robust learned image compression.
\newblock \emph{arXiv preprint arXiv:2401.11902}, 2024.

\bibitem[Srinivasan et~al.(2021)Srinivasan, Rohrer, Marban, M{\"u}ller, Samek, and Nakajima]{srinivasan2021robustifying}
Srinivasan, V., Rohrer, C., Marban, A., M{\"u}ller, K.-R., Samek, W., and Nakajima, S.
\newblock Robustifying models against adversarial attacks by langevin dynamics.
\newblock \emph{Neural Networks}, 137:\penalty0 1--17, 2021.

\bibitem[Szegedy et~al.(2014)Szegedy, Zaremba, Sutskever, Bruna, Erhan, Goodfellow, and Fergus]{szegedy2013intriguing}
Szegedy, C., Zaremba, W., Sutskever, I., Bruna, J., Erhan, D., Goodfellow, I., and Fergus, R.
\newblock Intriguing properties of neural networks.
\newblock \emph{International Conference on Learning Representations}, 2014.

\bibitem[Ulyanov et~al.(2018)Ulyanov, Vedaldi, and Lempitsky]{ulyanov2018deep}
Ulyanov, D., Vedaldi, A., and Lempitsky, V.
\newblock Deep image prior.
\newblock In \emph{Proceedings of the IEEE conference on computer vision and pattern recognition}, pp.\  9446--9454, 2018.

\bibitem[Wang et~al.(2022)Wang, Lyu, Lin, Dai, and Fu]{wang2022guided}
Wang, J., Lyu, Z., Lin, D., Dai, B., and Fu, H.
\newblock Guided diffusion model for adversarial purification.
\newblock \emph{arXiv preprint arXiv:2205.14969}, 2022.

\bibitem[Wang et~al.(2023)Wang, Pang, Du, Lin, Liu, and Yan]{wang2023better}
Wang, Z., Pang, T., Du, C., Lin, M., Liu, W., and Yan, S.
\newblock Better diffusion models further improve adversarial training.
\newblock \emph{International conference on machine learning}, 2023.

\bibitem[Wong et~al.(2020)Wong, Rice, and Kolter]{wong2020fast}
Wong, E., Rice, L., and Kolter, J.~Z.
\newblock Fast is better than free: Revisiting adversarial training.
\newblock \emph{arXiv preprint arXiv:2001.03994}, 2020.

\bibitem[Yang et~al.(2019)Yang, Zhang, Katabi, and Xu]{yang2019me}
Yang, Y., Zhang, G., Katabi, D., and Xu, Z.
\newblock Me-net: Towards effective adversarial robustness with matrix estimation.
\newblock \emph{International Conference on Machine Learning}, 2019.

\bibitem[Yoon et~al.(2021)Yoon, Hwang, and Lee]{yoon2021adversarial}
Yoon, J., Hwang, S.~J., and Lee, J.
\newblock Adversarial purification with score-based generative models.
\newblock In \emph{International Conference on Machine Learning}, pp.\  12062--12072. PMLR, 2021.

\bibitem[Zagoruyko \& Komodakis(2016)Zagoruyko and Komodakis]{zagoruyko2016wide}
Zagoruyko, S. and Komodakis, N.
\newblock Wide residual networks.
\newblock In \emph{Procedings of the British Machine Vision Conference 2016}. British Machine Vision Association, 2016.

\bibitem[Zhang et~al.(2020)Zhang, Zhu, Niu, Han, Sugiyama, and Kankanhalli]{zhang2020geometry}
Zhang, J., Zhu, J., Niu, G., Han, B., Sugiyama, M., and Kankanhalli, M.
\newblock Geometry-aware instance-reweighted adversarial training.
\newblock In \emph{International Conference on Learning Representations}, 2020.

\bibitem[Zhang et~al.(2024)Zhang, Li, Chen, Guo, and Cheng]{anonymous2023classifier}
Zhang, M., Li, J., Chen, W., Guo, J., and Cheng, X.
\newblock Classifier guidance enhances diffusion-based adversarial purification by preserving predictive information, 2024.
\newblock URL \url{https://openreview.net/forum?id=qvLPtx52ZR}.

\bibitem[Zhao et~al.(2016)Zhao, Zhou, Xie, Zhang, and Cichocki]{zhao2016tensor}
Zhao, Q., Zhou, G., Xie, S., Zhang, L., and Cichocki, A.
\newblock Tensor ring decomposition.
\newblock \emph{arXiv preprint arXiv:1606.05535}, 2016.

\end{thebibliography}
\bibliographystyle{icml2025}

\newpage
\appendix
\clearpage
\onecolumn
\section*{Appendix}

\section{Influence of different sampling methods}
\label{app:avgpool}
To support our hypothesis of using the average pooling, we test it with stride sampling, which selects pixels with constant strides.
In principle, the stride sampling would not change the distribution of perturbations. Therefore, it serves as a baseline to compare the influence of distributions.

\begin{figure}[h]
    \centering
    \subfigure[MoG with Avg Pooling]{%
        \includegraphics[width=0.47\linewidth]{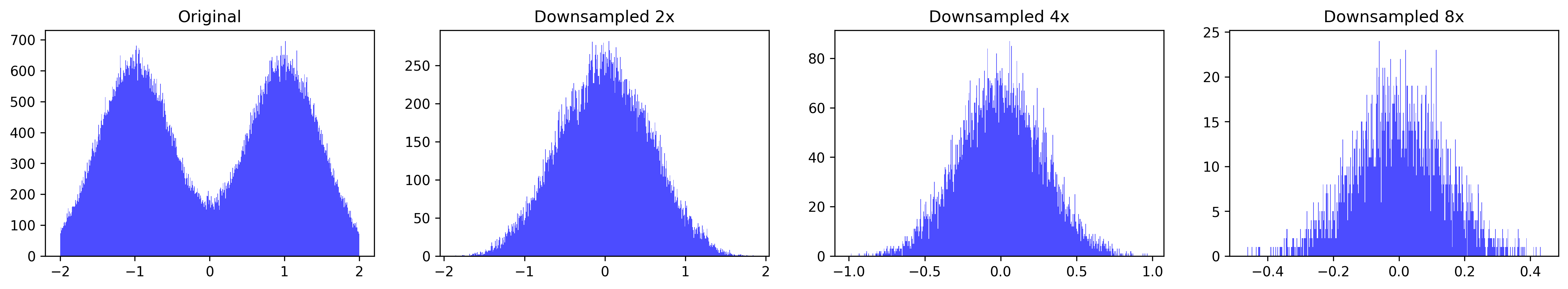} % Replace with your image
    }
    \subfigure[MoG with Stride Sampling]{%
        \includegraphics[width=0.47\linewidth]{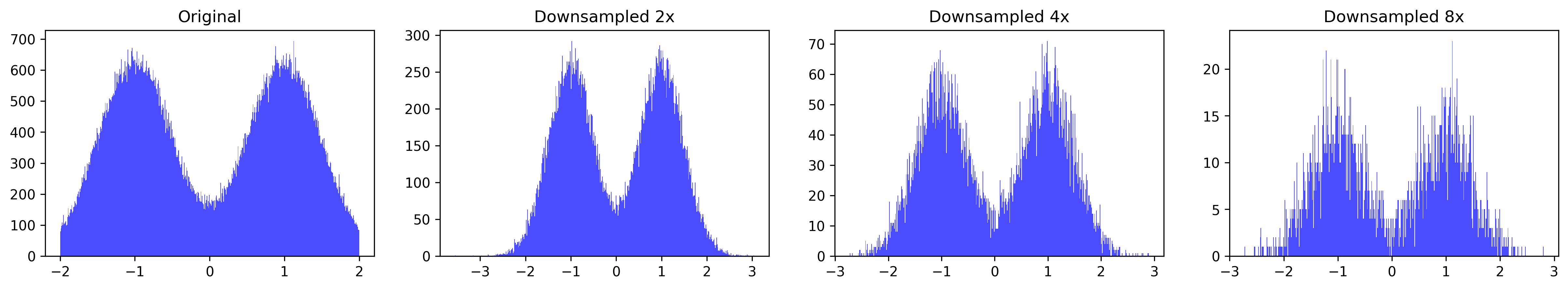} % Replace with your image
    }

    \subfigure[Beta with Avg Pooling]{%
        \includegraphics[width=0.47\linewidth]{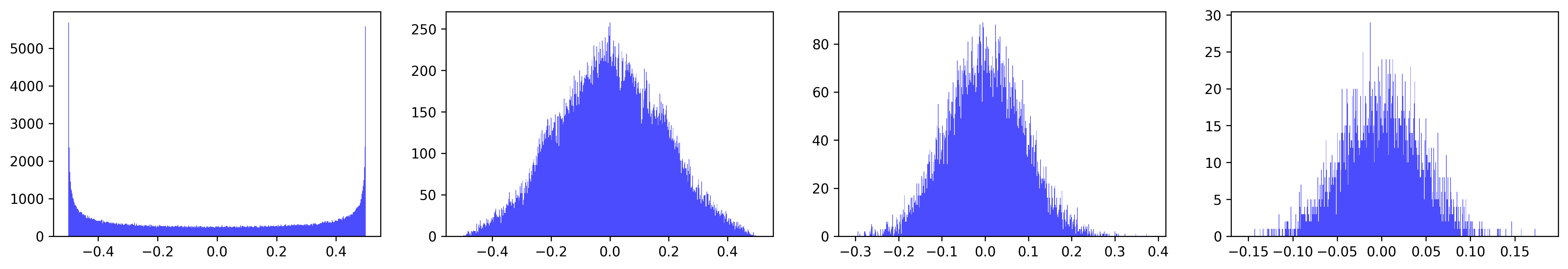} % Replace with your image
    }
    \subfigure[Beta with Stride Sampling]{%
        \includegraphics[width=0.47\linewidth]{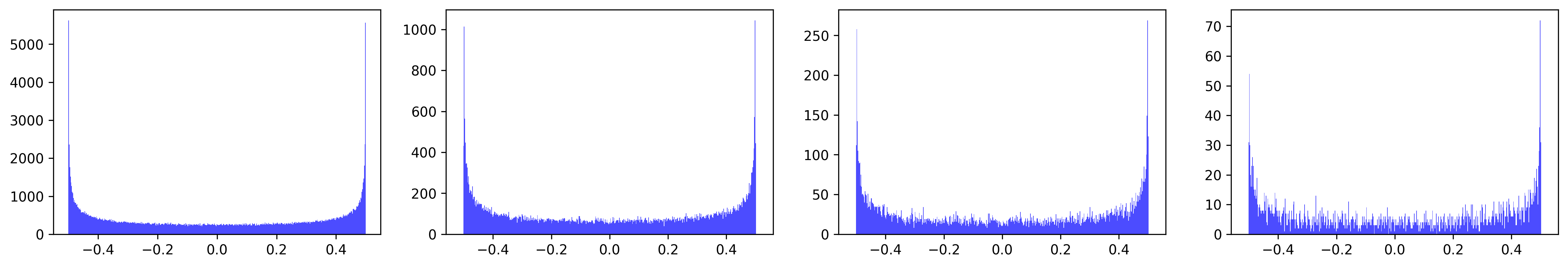} % Replace with your image
    }

    \subfigure[Uniform with Avg Pooling]{%
        \includegraphics[width=0.47\linewidth]{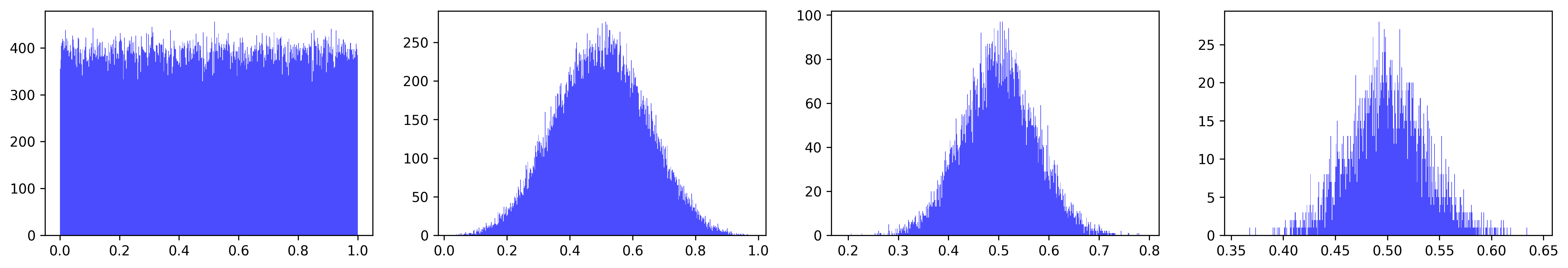} % Replace with your image
    }
    \subfigure[Uniform with Stride Sampling]{%
        \includegraphics[width=0.47\linewidth]{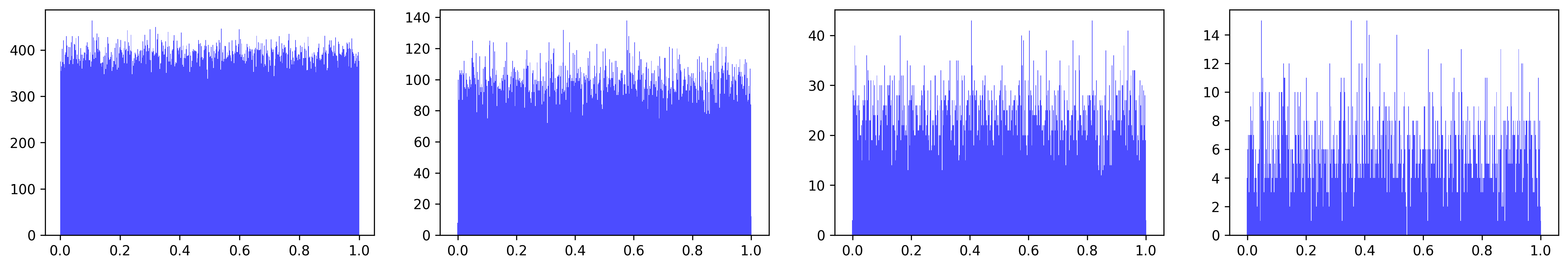} % Replace with your image
    }
    \caption{Histogram figures of noises under different sampling methods.}
    \label{fig:app-sampling}
\end{figure}

\begin{figure}[h]
    \centering
    \subfigure[PSNR]{%
    \includegraphics[width=0.4\linewidth]{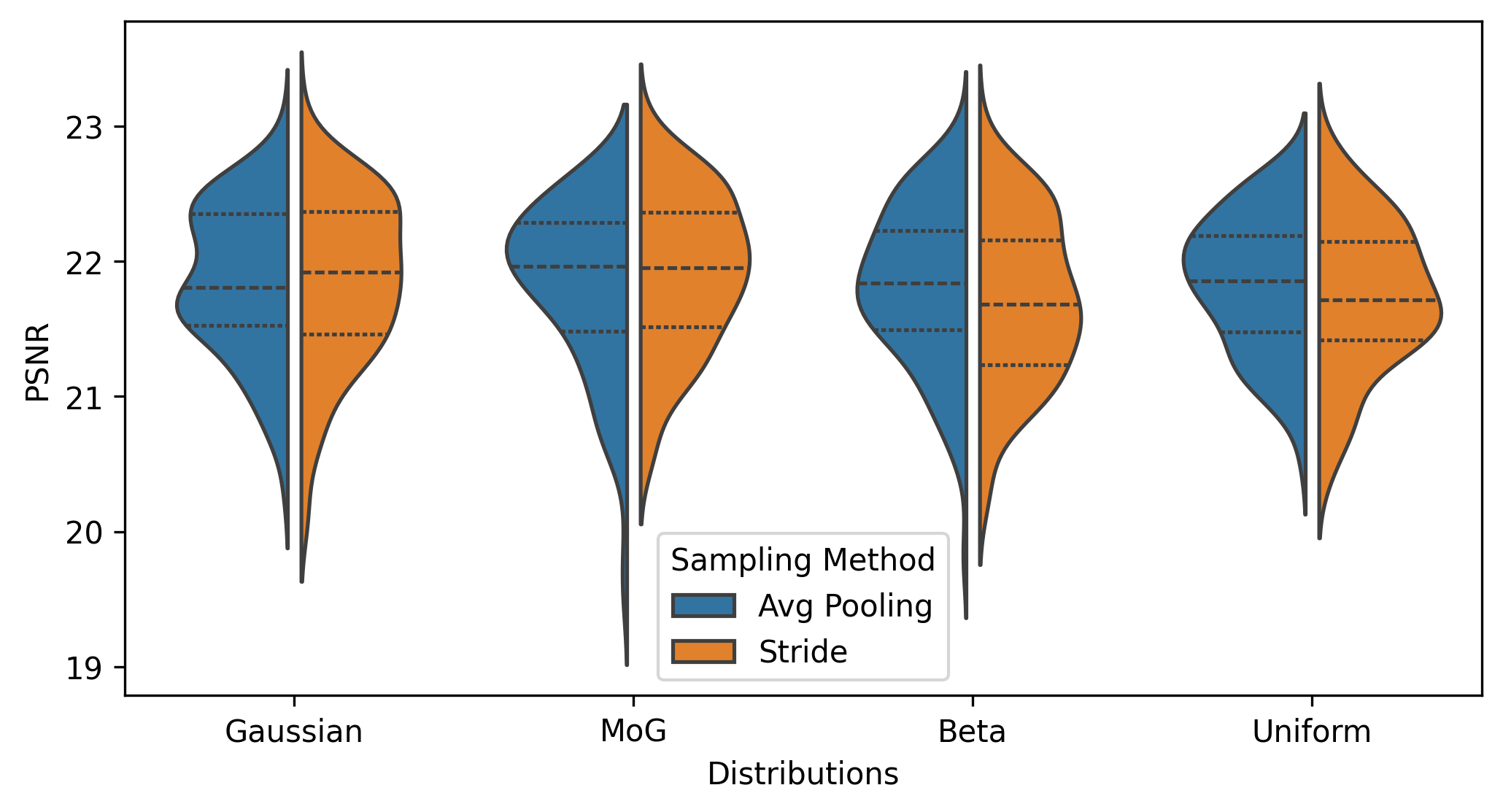} % Replace with your image
    }\hspace{2em}
    \subfigure[SSIM]{%
    \includegraphics[width=0.4\linewidth]{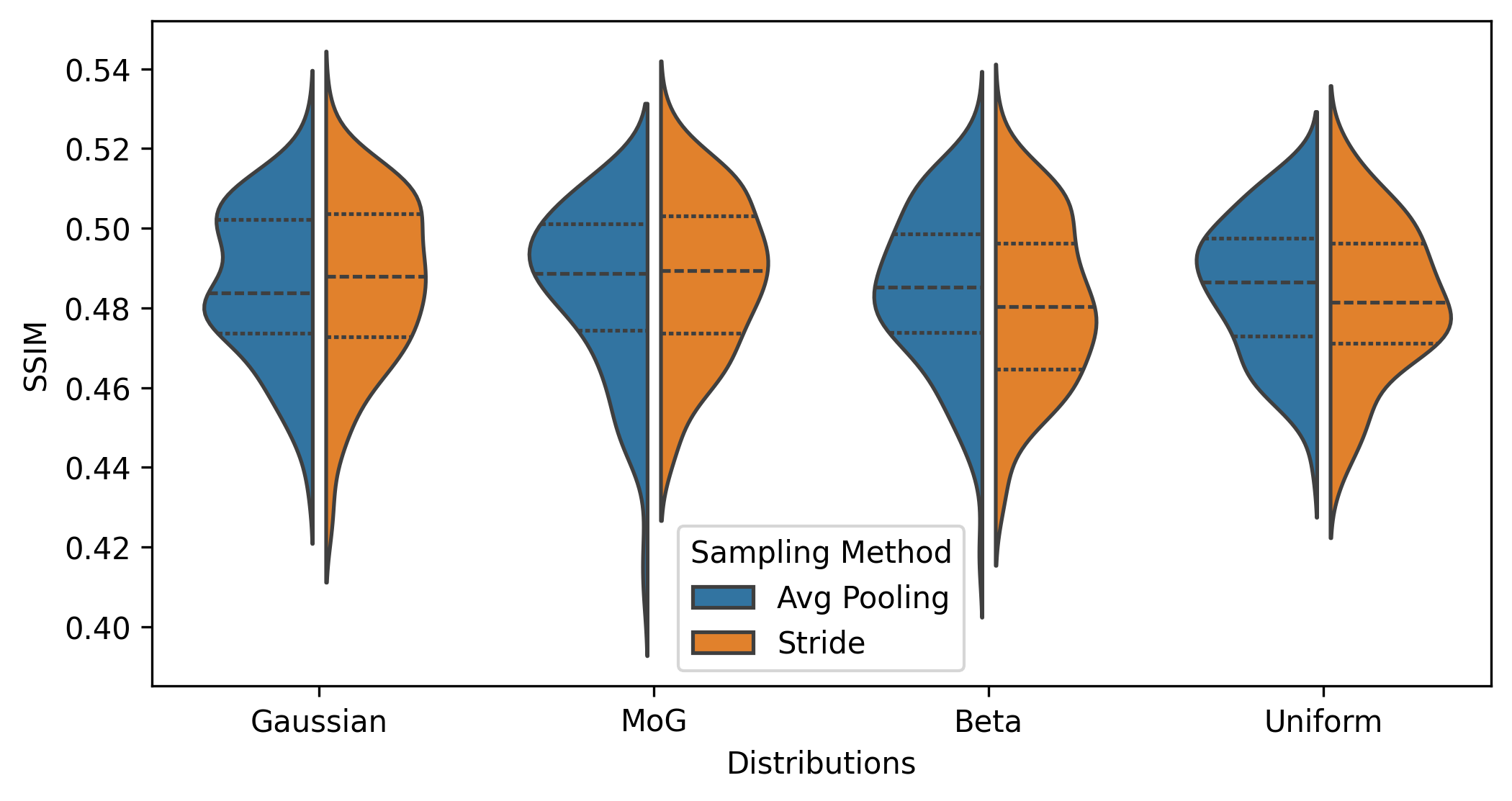} % Replace with your image
    }
    \caption{Violin plot of denoising results using different sampling methods. (a) PSNR results. (b) SSIM results.}
    \label{fig:app-denoising}
\end{figure}

We test four types of noise distributions: (1) Gaussian $\mathcal{N}(0, 0.3^2)$, (2) Mixture of Gaussian (MoG), $0.5\cdot\mathcal{N}(-1.0, 0.5^2) + 0.5\cdot\mathcal{N}(1.0, 0.5^2)$, (3) Beta distribution, $\text{Beta}(0.5, 0.5) - 0.5$, and (4) Uniform distribution, $\text{Uniform}(-0.5, 0.5)$. For MoG, Beta and uniform noises, we scale them to have the same signal-to-noise ratio with the Gaussian distribution. We add the noises on the Girl image \cite{loeschcke2024coarse} with resolution $1024 \times 1024$. First, we show the noise distributions in \Cref{fig:app-sampling}. As can be seen, the Avg Pooling strategy transforms the non-Gaussian noises into Gaussian-like noises, while the Stride sampling would not. Second, we run the PuTT algorithm with different sampling methods for 100 times. The violin plot of denoising results are shown in \Cref{fig:app-denoising}.
In Gaussian distribution, the Stride sampling is better than AvgPooling. While for non-Gaussian noises, the AvgPooling is more robust and better than Stride.
The denoising results indicate that the average pooling can handle different types of noises, which is consistent with our hypothesis. However, as we introduced, this might not be enough, since we need to deal with the original image and noises in the final stage.

\section{Tensor network decomposition}\label{app:TN}

\subsection{Matrix Product Operators}
A matrix product operator (MPO) \cite{mcculloch2008infinite,hubig2017generic} is the TN representation of
a linear operator acting on a TT format, which makes it highly efficient to handle large operators.
Namely, a linear operator
\(
\vectcal{A}: \Real^{I_1 \times \ldots \times I_D} \to \Real^{J_1 \times \ldots \times J_D} \,.
\)
Namely, if $\vectcal{Y}=\vectcal{A}\vectcal{X}$, then each entry of $\vectcal{Y}$ is given as
\[
{y}_{\mathbf{i}}= \sum_{i_1=1}^{I_1} \cdots \sum_{i_D=1}^{I_D} \vect{A}^1_{j_1,i_1} \vect{A}^2_{j_2,i_2} \ldots \vect{A}^D_{j_D,i_D} \vect{X}^1_{i_1} \vect{X}^2_{i_2} \ldots \vect{X}^D_{i_D}  \,,
\]

\subsection{Prolongation Operator}\label{app:TN.P}
This work uses a specific MPO, known as the
prolongation operator $\vectcal{P}_d$ \cite{lubasch2018multigrid}, to upsample a QTT format
of an image from resolution $d-1$ to $d$.

Consider a one-dimensional vector $\vect{x}_d \in \Real^{2^d}$.
The matrix $\vect{P}_{2^d \to 2^{d+1}}$ upsamples $\vect{x}_d$ to $\vect{x}_{d+1}$ by linear interpolation between adjacent points.
For example, for $d=2$,
\[
\vect{P}_{4 \to 8} =
\begin{bmatrix}
1 & 0 & 0 & 0 \\
0.5 & 0.5 & 0 & 0 \\
0 & 1 & 0 & 0 \\
0 & 0.5 & 0.5 & 0 \\
0 & 0 & 1 & 0 \\
0 & 0 & 0.5 & 0.5 \\
0 & 0 & 0 & 1 \\
0 & 0 & 0 & 0.5
\end{bmatrix}
\]
The matrix $\vect{P}_{2^d \to 2^{d+1}}$ can be written
as an MPO $\vectcal{P}_{d+1}$ entry-wise
\[
p_{j_1,\ldots, j_d, i_1, \ldots, i_{d+1}} = \vect{P}^1_{j_1,i_1} \ldots \vect{P}^{d}_{j_d,i_d} \vect{P}^{d+1}_{i_{d+1}} \,.
\]
The entries are given explicitly \cite{lubasch2018multigrid} as
\begin{align}
    \vect{P}^l_{1,1}(1,1) &= \vect{P}^l_{2,2}(1,1) = \vect{P}^l_{2,1}(1,2) = \vect{P}^l_{1,2}(2,2) = 1, \forall l \in [d] \notag \\
    \vect{P}^{d+1}_{1}(1) &= 1 \,, \vect{P}^{d+1}_{2}(1) = \vect{P}^{d+1}_{2}(2) = 0.5 \notag \,,
\end{align}
and other entries are zero.

The prolongation operator described above applies to the QTT format of
one-dimensional vectors. In general, this operator is the tensor product of the one-dimensional operators on each dimension: $\vectcal{P}_d^{(2)}=\vectcal{P}_d \otimes \vectcal{P}_d$ for 2-dimensions (images)
and $\vectcal{P}_d^{(3)}=\vectcal{P}_d \otimes \vectcal{P}_d \otimes \vectcal{P}_d$ for 3-dimensions (3D objects).
For simplicity, since this work concerns only images,
the superscript is omitted, denoting
the prolongation operator as $\vectcal{P}_d$.

Ultimately, for a resolution $d$ image $\vect{x}_d$, and
$\vectcal{X}_d = \quant(\vect{x}_d)$,
the upsampled image is resolution $d+1$, given as $\text{P}_d(\vect{x}_{d}) = \quant^{-1}(\vectcal{P}_d\vectcal{X}_d)$,
where the linear function $\text{P}_d(\cdot)$ acts on the image level.

\subsection{Recap of PuTT \cite{loeschcke2024coarse}}
\label{app:TN.putt}

A $2^D \times 2^D$ image, denoted as $\vect{x}_D$, can be quantized in to a $D$th
order tensor $\vectcal{X}_D=\text{Q}(\vect{x}_D)$.
Firstly, $\vect{x}_D$ is downsampled by average pooling to $\vect{x}_{D-l}$, correspondingly possesing a quantization $\vectcal{X}_{D-l}$.
Then, $D-l$ QTT cores of ${X}_{D-l}$ can be optimized by backpropagation, returning $\vectcal{Y}_{D-l}$.
The QTT cores of next resolution $\vectcal{X}_{D-l+1}$
can be optimized similarly, initialized by the prologation $\vectcal{P}_{D-l+1} (\vect{y}_{D-l})$.
Repeat the process until the original resolution. \cite{loeschcke2024coarse} demonstrates impressive reconstruction
capability of PuTT thanks to the QTT structure and coarse-to-fine approach.
The pseudocode is given in \cref{alg:putt}.

\begin{algorithm}[H]
    \caption{PuTT \cite{loeschcke2024coarse}}
    \label{alg:putt}
    \begin{algorithmic}
        \STATE \textbf{Input:} Image $\vect{x}_D$, number of iterations $T$,
        upsampling iterations $(t_1, \ldots, t_l)$.
        \STATE \textbf{Output:} TT reconstruction $\vect{y}_D = \text{PuTT}(\vect{x}_D)$.
        \STATE $d\gets D-l \,, \vect{x}_{d}\gets\avgpool(\vect{x}_D) \,, \vectcal{X}_{d}\gets \quant(\vect{x}_d)$
        \FOR{$t=1\to T$}
        \IF{$t \in (t_1, \ldots, t_l)$}
            \STATE $d \gets d+1$
            \STATE $\vect{x}_{d} \gets \avgpool(\vect{x}_D)$
            \STATE $\vectcal{X}_{d} \gets \quant(\vect{x}_d)$
        \ENDIF
        \STATE Loss $\ell \gets \text{MSE}(\vectcal{Y}_d-\vectcal{X}_d)$
        \STATE Update QTT cores $\vectcal{Y}_d$ by backpropagation
        \ENDFOR
        \STATE \textbf{return} $\vect{y}_D = \quant^{-1}(\vectcal{Y}_D)$
    \end{algorithmic}
\end{algorithm}

However, while PuTT aims to obtain better initialization by downsampling for better optimization and reconstruction, it does not account for adversarial examples or analyze the impact of downsampling on perturbations. Additionally, PuTT also minimizes the reconstruction loss on the input image, which inevitably results in the reconstruction of the perturbations. In contrast, we focus on the perturbations and propose a new optimization process introduced in the next section, aiming to reconstruct clean examples.

\section{More details of experimental settings}
\label{app:settings}

\subsection{Implementation details of adversarial attacks}
\textbf{AutoAttack} \quad We evaluate our method of defending against AutoAttack \citep{croce2020reliable} and compare with the state-of-the-art methods as listed RobustBench benchmark (https://robustbench.github.io). For a comprehensive evaluation, we conduct experiments under all adversarial attack settings. Specifically, we set $\epsilon=8/255$ and $\epsilon=0.5/1.0$ for AutoAttack $l_{\inf}$ and AutoAttack $l_2$ threats on CIFAR-10. On CIFAR-100, we set $\epsilon=8/255$ for AutoAttack $l_{\inf}$. On ImageNet, we set $\epsilon=4/255$ for AutoAttack $l_{\inf}$.

\textbf{PGD+EOT} \quad We evaluate our method of defending against PGD+EOT \citep{madry2018towards,athalye2018synthesizing} and present the comparisons of AT methods, AP methods, and our method. Following the guidelines of \citet{lee2023robust}, we set $\epsilon=8/255$ for PGD+EOT $l_{\inf}$ threats on CIFAR-10, where the update iterations of PGD is 200 with 20 EOT samples.

\subsection{Implementation details of our method}
For CIFAR-10, CIFAR-100 with resolution $32 \times 32$ and ImageNet with resolution $224 \times 224$, we first upsample them into resolution $2^D \times 2^D$ image $x_{D}$. Based on the initial experimental results, we set  $D = 8$, $l = 1$, $\alpha=0.1$, $\eta=0.1$ and $N=1$ for the following experiments. The table results presented in the paper are conducted under these hyperparameters. This trick creates a large enough image to downsample until the perturbations are well mixed into Gaussian noise. Furthermore, without this initial step, the semantic information can become almost indistinguishable after several downsampling steps, especially for low-resolution images.
For example, if a $32\times 32$ image is reduced with the factor of 8, the resolution $4\times 4$ image is of poor quality. Additionally, to more clearly observe the denoising effects in visualization results, we upsample the images to resolution $D=11$ with $\alpha=0.05$, $\eta=0.1$ and $N=3$ for the experiments in \Cref{fig:visual}, and comparisons in different downsampled images in \Cref{fig:distribution}. The code will be available upon acceptance, with more details provided in the configuration files.

\subsection{Implementation details of evaluation metrics}

We evaluate the performance of defense methods using multiple metrics: Standard accuracy and robust accuracy \citep{szegedy2013intriguing} on classification tasks. For denoising tasks, we measure the Normalized Root Mean Squared Error \citep[NRMSE,][]{botchkarev2018performance}, Structural Similarity Index Measure \citep[SSIM,][]{hore2010image}, Peak Signal-to-Noise Ratio (PSNR) metrics between a reference image $\vect{x}$ and its reconstruction $\vect{y}$, where pixel values are in $[0,1]$.
In denoising and reconstruction tasks, a lower NRMSE, a higher SSIM, and a higher PSNR generally indicate better performance.

% \textbf{Normalized Root Mean Squared Error}
% \[
% \text{NRMSE}(\vect{x}, \vect{y}) = \frac{\norm{\vect{x}-\vect{y}}_2}{\norm{\vect{x}}_2} \,.
% \]

% \textbf{Structural Similarity Index Measure \cite{hore2010image}} \quad
% \[
% \text{SSIM}(\vect{x}, \vect{y}) = \frac{(2\mu_x \mu_y + C_1)(2\sigma_{xy} + C_2)}
% {(\mu_x^2 + \mu_y^2 + C_1)(\sigma_x^2 + \sigma_y^2 + C_2)}
% \]

% where:
% \begin{itemize}
%     \item \( \mu_x \) and \( \mu_y \) are the mean pixel values of images \( \vect{x} \) and \( \vect{y} \).
%     \item \( \sigma_x^2 \) and \( \sigma_y^2 \) are the variances of \( \vect{x} \) and \( \vect{y} \).
%     \item \( \sigma_{xy} \) is the covariance between \( \vect{x} \) and \( \vect{y} \).
%     \item \( C_1 \) and \( C_2 \) are small constants to stabilize the division.
% \end{itemize}

% \textbf{Peak Signal-to-Noise Ratio} \quad
% \[
% \text{PSNR}(\vect{x}, \vect{y}) = 10 \log_{10} \left( \frac{1}{\text{MSE}(\vect{x}, \vect{y})} \right) \,.
% \]

\section{Inference time cost}
\begin{table}[htbp]
\centering
\caption{Inference time}
\vskip 0.15in
\label{tab:inference-time}
\begin{tabular}{cccc}
\toprule
Methods & CIFAR-10 & CIFAR-100 & ImageNet \\
\midrule
AT & 0.002 s & 0.002 s & 0.005 s \\
DM-based AP & 1.49 s & 1.50 s & 5.11 s \\
Ours & 12.39 s & 11.81 s & 14.10 s \\
% Ours & 11.30/12.39 s & 11.27/11.81 s & 14.10 s \\
\bottomrule
\bottomrule
\end{tabular}
\vskip -0.1in
\end{table}

\Cref{tab:inference-time} shows the inference time of different methods on CIFAR-10, CIFAR-100, and ImageNet, which is measured on a single image. Specifically, AP method purifies CIFAR data at a resolution of $32 \times 32$ and ImageNet data at $256 \times 256$, whereas our method operates at a resolution of $256 \times 256$ across all datasets. As a form of Test-Time Training, our method inevitably increases inference cost. In a comparison at the same resolution of ImageNet, the AP method require 5.11 seconds, whereas our method takes 14.10 seconds. This overhead stems from the inherent limitations of iterative optimization processes and affects their practicality in real-world applications. We leave the study of integrating our TN-based AP technique with more advanced and faster optimization strategies for future research.

\section{Zero-shot adversarial defense}

AT and AP methods depend heavily on external training dataset, overlooking the
potential internal priors in the input itself.
Among adversarial defense techniques, untrained neural networks such
as deep image prior (DIP) \cite{ulyanov2018deep} and masked autoencoder (MAE) \cite{he2022masked} have been
utilized to avoid the need of extra training data \cite{dipdefend2020,dai2022deep,lyu2023maedefense}.
However, although such deep learning models achieve high-quality reconstruction results,
they have been shown to be susceptible to revive also the adversarial noise.
This section compares two representative untrained models DIP and MAE.

\begin{table}[h]
    \caption{Comparison with untrained neural networks against AutoAttack $l_\infty$ threat ($\epsilon=8/255$) on CIFAR-10.}
    \vskip 0.15in
    \label{tab:trainingfree}
    \begin{center}
    \begin{tabular}{ccccc}
    \toprule
    Defense method & Acc. & NRMSE & SSIM & PSNR  \\
    \midrule
    \multicolumn{5}{l}{Clean examples} \\
    \midrule
    DIP & 90.43 & 0.0464 & 0.9565 & 32.13 \\
    MAE & 88.28 & 0.0847 & 0.8842 & 26.90  \\
    Ours & 82.23 & 0.0644 & 0.9203 & 29.06 \\
    \midrule
    \multicolumn{5}{l}{Adversarial examples} \\
    \midrule
    DIP & 38.28 & 0.0451 & 0.9467 & 32.53  \\
    MAE & 1.56 & 0.0914 & 0.8472 & 26.24  \\
    Ours & 55.27 & 0.0748 & 0.8707 & 27.77 \\
    \bottomrule
    \bottomrule
    \end{tabular}
    \end{center}
    \vskip -0.1in
\end{table}

\cref{tab:trainingfree} shows that although DIP and MAE have achieved remarkable
standard accuracy and reconstruction quality, they deteriorate significantly
under attack.

\section{More discussion}
As we all know, the adversarial challenge of attack and defense is endless. This contradiction arises from the fundamental difference between adversarial attacks and defenses. Attacks are inherently destructive, whereas defenses are protective. This adversarial relationship places the attacker in an active position, while the defender remains passive. As a result, attackers can continually explore new attack strategies against a fixed model to degrade its predictive performance, ultimately leading to the failure of conventional defenses.
The introduction of TNP has the potential to address this issue. As a model-free technique, TNP generates tensor representations solely based on the input information. These representations dynamically change with each input, preventing attackers from exploiting a fixed model to generate effective adversarial examples. This defensive mechanism allows TNP to maintain a more proactive stance in the ongoing competition between adversarial attacks and defenses.

\section{Histogram and kernel density estimation results}
\label{app:distribution}
\begin{figure}[ht]
\vskip 0.2in
    \centering
    \includegraphics[width=\linewidth]{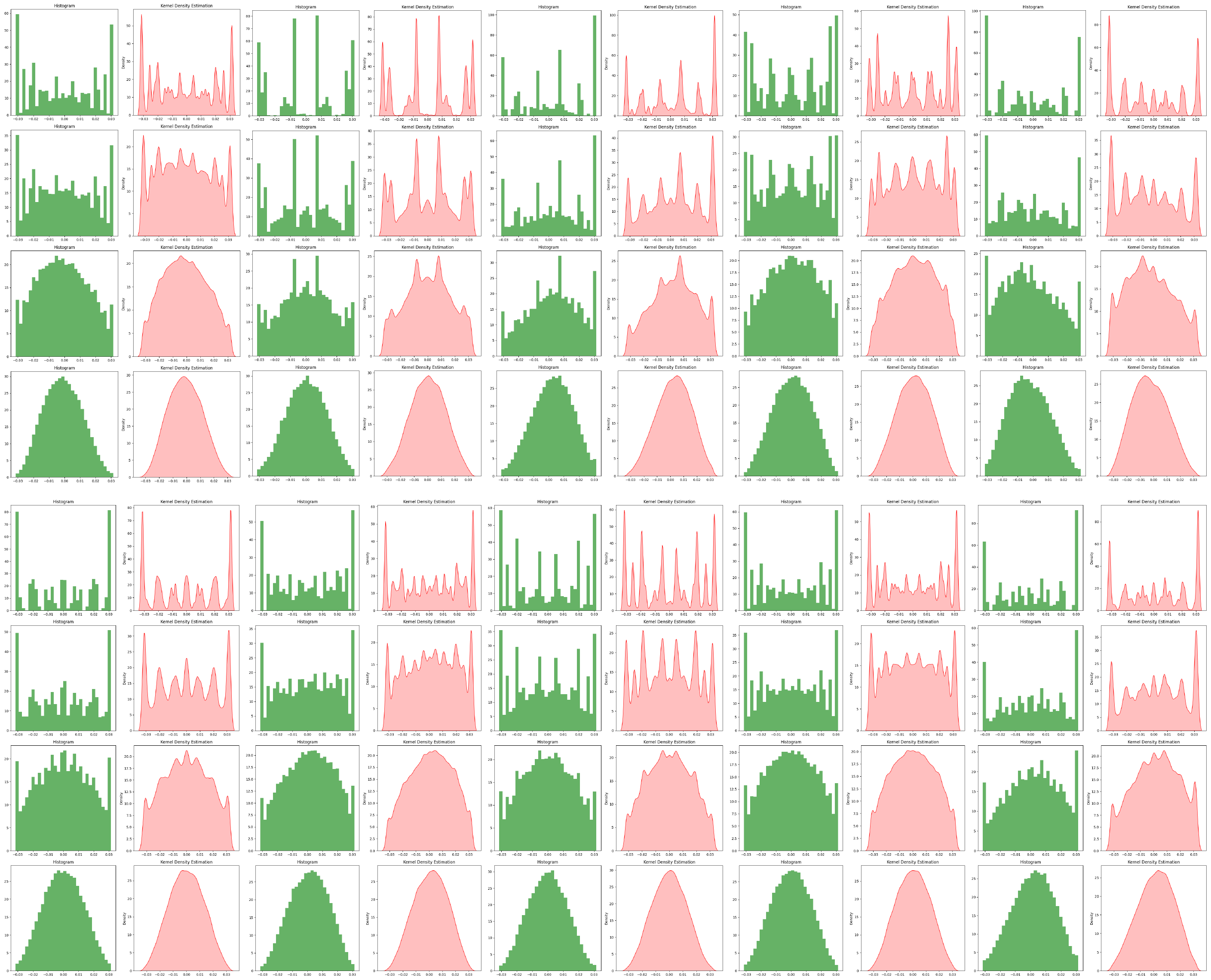}
    \caption{The histogram and kernel density estimation of adversarial perturbations in the downsampled images.}
    \label{fig:app-distribution}
    \vskip -0.1in
\end{figure}

\Cref{fig:app-distribution} shows the histogram and kernel density estimation of adversarial perturbations on 10 images. The distribution of those perturbations progressively aligns with that of Gaussian noise as the downsampling process progresses.

\section{Visualization}
\begin{figure}[ht]
\vskip 0.2in
    \centering
    \includegraphics[width=0.9\linewidth]{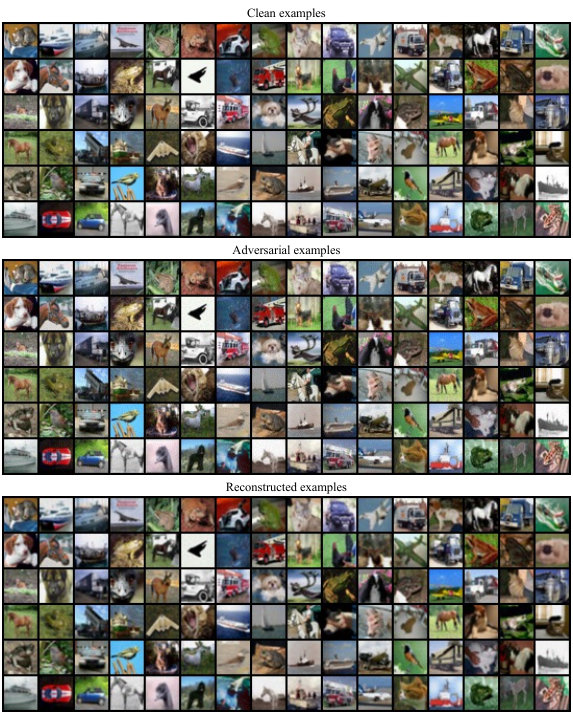}
    \caption{Clean examples (Top), adversarial examples (Middle) and reconstructed examples (Bottom) of CIFAR-10.}
    \label{fig:app-visual}
    \vskip -0.1in
\end{figure}

\end{document}